\pdfoutput=1
\documentclass{article}
\usepackage{iclr2027_conference,times}

\usepackage{amsmath,amsfonts,bm}

\def\eqref#1{equation~\ref{#1}}

\def\1{\bm{1}}

\DeclareMathAlphabet{\mathsfit}{\encodingdefault}{\sfdefault}{m}{sl}
\SetMathAlphabet{\mathsfit}{bold}{\encodingdefault}{\sfdefault}{bx}{n}

\newcommand{\E}{\mathbb{E}}

\usepackage{hyperref}
\hypersetup{hidelinks, pdftitle={Tokenizer Myths and Where To Find Them}, pdfauthor={Rylan Schaeffer, Sanmi Koyejo}}
\usepackage{url}
\usepackage{amsmath}
\usepackage{amssymb}
\usepackage{amsthm}
\usepackage{graphicx}
\usepackage{booktabs}
\usepackage{array}
\usepackage{multirow}
\usepackage{xcolor}
\usepackage{cleveref}
\usepackage{enumitem}
\usepackage{epigraph}

\newtheorem{theorem}{Theorem}
\newtheorem{proposition}{Proposition}

\theoremstyle{definition}

\theoremstyle{plain}



\newcommand{\bpb}{\operatorname{BPB}}
\newcommand{\nll}{\operatorname{NLL}}

\title{Tokenizer Myths and Where To Find Them}

\author{Rylan Schaeffer \\
Stanford Computer Science \\
\texttt{rschaef@cs.stanford.edu}
\And
Sanmi Koyejo \\
Stanford Computer Science \\
\texttt{sanmi@cs.stanford.edu}}

\iclrfinalcopy

\begin{document}

\maketitle
\lhead{Preprint}

\begin{abstract}
Tokenization is a fundamental component of modern language models: before a model sees data, its tokenizer has already determined how the data appears.
We identify and challenge three myths in the tokenization research literature.
Myth \#1: Bits-Per-Byte (BPB) is agnostic to the tokenizer and/or the vocabulary size:
BPB is tokens per byte times bits per token, and the two factors do not cancel; using untrained models, we show that BPB prefers some tokenizers over others: the model has learned nothing, but the tokenizer--model pair has.
Myth \#2: lower BPB means a better model:
On untrained models, BPB strongly prefers some vocabulary sizes, even though accuracy remains at chance, and on real models trained on the same data, BPB correlates weakly with downstream performance.
Myth \#3: different ``tokenizer-agnostic'' metrics yield consistent results:
The ``unigram-normalized loss'' is not tokenizer-agnostic; its reported correlation of $0.989$ with BPB comes from a confounder and falls to $0.057$ once fixed; from the same data the two metrics select different vocabularies at each model size.
We present a tokenizer trilemma: no metric can score only the true distribution best, give the same predictions the same score, and tie all untrained models, so choosing a metric is choosing a research question.
We conclude with a caution for reading the tokenization literature and a conjecture for future work.
\end{abstract}

\section{Introduction}
\label{sec:intro}

Tokenizers shape what language models see: a tokenizer cuts raw text into the units that a model reads and writes.
Tokenization research asks how this choice affects a wide range of model properties, such as the model's cost in different languages \citep{petrov2023unfairness, ahia2023allcost}, its compute at inference \citep{liu2025superbpe, pagnoni2024blt}, the amount of training data it needs \citep{hwang2026dynamic}, its spelling and its robustness to noise \citep{xue2022byt5}, its arithmetic \citep{singh2024tokenizationcounts}, its under-trained tokens \citep{land2024magikarp}, and the vocabulary size that suits its parameter count \citep{tao2024scalingvocab}, among many others \citep{cognetta2026tokenizationsurvey}.
Many of these studies compare language models that read the same text through different tokenizers: models with different vocabulary sizes \citep{tao2024scalingvocab, meister2026tokeval}, models whose vocabularies come from different algorithms \citep{ali2024tokenizerchoice, schmidt2024morethancompression, liu2025superbpe, altintas2026toksuite}, and byte-level models against token-level ones \citep{xue2022byt5, yu2023megabyte, wang2024mambabytetokenfreeselectivestate, slagle2024spacebyte, pagnoni2024blt, hwang2026dynamic}.

These comparisons need metrics to quantify the effects of tokenization, and whatever metric is chosen must compare tokenizers on a level playing field. 
One standard metric, the token-averaged negative log likelihood, is unfair because it divides the total loss by the number of tokens, and different tokenizers cut the same text into different numbers of tokens.
The common remedy is Bits-Per-Byte ($\bpb$), which divides the same total loss by the number of bytes.
A second remedy proposes a metric which divides each token's probability by the token's frequency \citep{roh2020unigramnorm}.
We examine three incorrect beliefs about these metrics that permeat the literature.

\begin{figure}[t!]
\centering
\includegraphics[width=\textwidth]{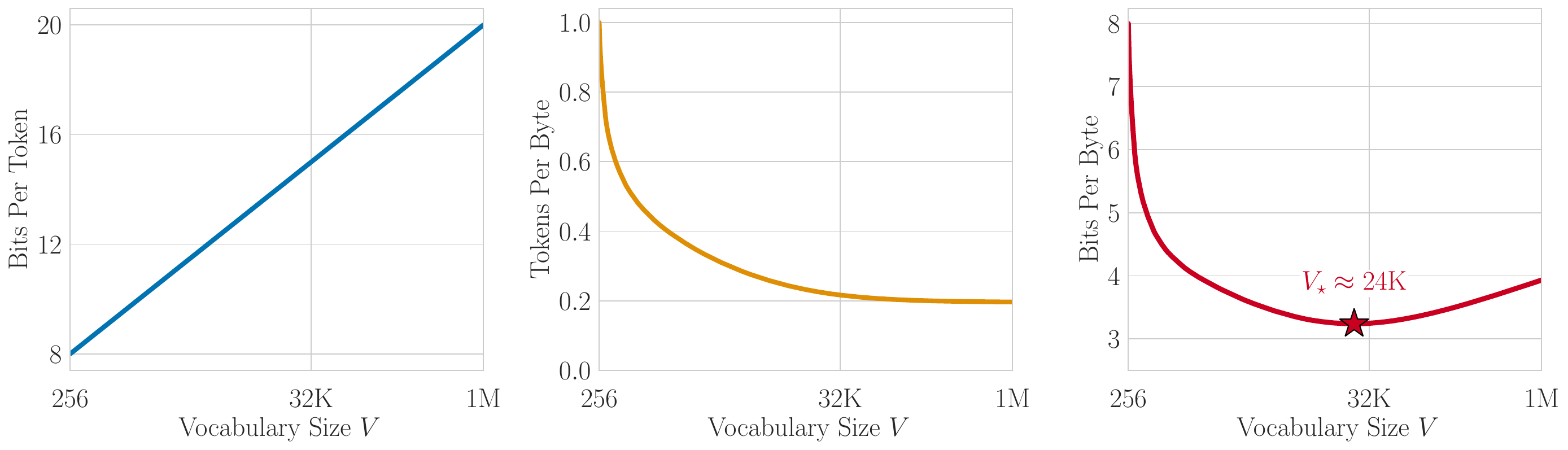}
\caption{\textbf{Myth \#1: Bits-Per-Byte Is Agnostic to the Tokenizer and/or Its Vocabulary Size.}
For untrained models that output uniform distributions, BPB on held-out test data prefers certain vocabulary sizes over others (right). For uniform distributions, BPB is a product of two functions: the number of bits per token, which increases with vocabulary size (left), and the number of tokens per byte, which decreases with vocabulary size (center). The product has a minimum (right).
Corpus: FineWeb-Edu-dedup, English \citep{penedo2024finewebdatasetsdecantingweb}.
}
\label{fig:headline}
\end{figure}

The incorrect first belief is that $\bpb$ is agnostic to the tokenizer and its vocabulary size (Myth~\#1, \Cref{sec:tokenizer_agnostic}); it fails the same untrained-model test that the per-token loss also fails.
To make the point, we consider randomly initialized models which are approximately uniform under modern parameterizations such as $\mu$P (\Cref{app:random-init}).
$\bpb$ is tokens per byte times bits per token, and the two factors move in opposite directions as the vocabulary grows, which permits a $\bpb$ minimum (around $24$K on English web text; \citealp{penedo2024finewebdatasetsdecantingweb}).
Thus, a $\bpb$ comparison across tokenizers includes a head start from the tokenizer alone, even when the model has learned nothing.

The second incorrect belief is that a lower $\bpb$ means a better model (Myth~\#2, \Cref{sec:downstream}).
Untrained models on standard benchmarks exhibit $\bpb$ which moves by more than 6 bits per byte, even while accuracy stays at random chance.
Trained models from TokEval \citep{meister2026tokeval} similarly exhibit weak correlations between $\bpb$ and downstream metrics on most benchmarks we test.
One reason is structural:
the code length of a correct answer splits exactly into the cost of answering with one of the options, the cost of choosing the correct option, and the cost of spelling it out (\Cref{eq:answer-factorization}), whereas accuracy reads only the middle term.
On 49 released models from TokSuite \citep{altintas2026toksuite} and TokEval, a change of tokenizer moves mostly the first term, which accuracy ignores (\Cref{fig:decision-share}).
Several tokenizer studies have reported that $\bpb$ and task scores disagree \citep{liu2025superbpe, land2026mingram, tempus2026convextok, yavuz2026objective, meister2026tokeval, marathe2026tokenceiling}; our factorization gives a general reason.

The third incorrect belief is that different ``tokenizer-agnostic'' metrics yield consistent results; it fails when comparing $\bpb$ against the unigram-normalized loss (Myth~\#3, \Cref{sec:unigram_normalized_loss}; \citealp{roh2020unigramnorm, tao2024scalingvocab}).
The loss scores each model against its own tokenizer's unigram model, but it equals the difference between the model's $\bpb$ and that unigram model's $\bpb$, times bytes per token, so the per-token denominator returns (\Cref{eq:unigram-normalized-loss-bpb}).
\citet{tao2024scalingvocab} support the loss with its correlation of $0.989$ with $\bpb$ across their runs, but this high correlation is attributable to a confounder, and removing it reduces the correlation to $0.057$.
From the same data, the two metrics select different vocabularies at each model size.

No metric can be tokenizer-agnostic in the full sense (\Cref{sec:impossibility}).
One would want a metric under which only the true distribution scores best, the same predictions get the same score, and untrained models tie under different tokenizers. No metric has all three (\Cref{thm:trilemma}).
$\bpb$ keeps the first two properties and gives up the third, which is Myth~\#1.
The unigram-normalized loss gives up the second: it makes each tokenizer's unigram model score zero, but it scores the same predictions differently under different tokenizers.
Two metrics that give up different properties measure different things, so nothing forces them to agree; this is the disagreement of Myth~\#3.

\Cref{app:related_work} reviews related work.
We conclude with a caution for reading comparisons across tokenizers and a conjecture about discrimination and compression (\Cref{sec:recommendations}). 
\section{Myth \#1: Bits-Per-Byte Is Agnostic to the Tokenizer and/or the Vocabulary Size}
\label{sec:tokenizer_agnostic}

To fairly compare models that use different tokenizers, many publications advocate for \textbf{Bits-Per-Byte (BPB)}, arguing that $\bpb$ is agnostic to tokenization scheme and vocabulary size:

\begin{enumerate}
    \item \citet{karpathy2026autoresearch}: ``Bits per byte (BPB): vocab size-independent evaluation metric.''
    \item \citet{pagnoni2024blt}: ``[W]e instead report Bits-Per-Byte (BPB), a tokenizer independent version of perplexity.''
    \item \citet{gao2020pile}: ``Bits per byte is preferred over bits per character or perplexity [...] due to its invariance to different tokenization schemes [...]''
    \item \citet{liu2025superbpe}: ``[W]e evaluate language modeling performance using bits-per-byte (BPB), which normalizes the loss by the tokenizer’s encoding efficiency to fairly compare models with different tokenizers.''
    \item \citet{limisiewicz2026computeoptimal}: ``Both works use normalized negative log likelihood enabling a fair comparison across tokenizers.''
    \item \citet{meister2026tokeval}: ``We evaluate the resulting models on bits-per-byte (a tokenizer-agnostic version of perplexity) [...].''
    \item \citet{slagle2024spacebyte}: ``To fairly compare [...] we measure the cross-entropy of the test dataset in terms of bits-per-byte.''
    \item \citet{kallini2024mrt5}: ``[W]e compare models using bits-per-byte (BPB), a tokenizer-agnostic alternative to perplexity [...].''
\end{enumerate}

\textbf{This folk wisdom is wrong.} To see why, we consider token-averaged negative log-likelihood that the community agrees is not tokenizer-agnostic, and then show that BPB fails the same test.:

Fix an evaluation corpus of $D$ documents, where document $d$ contains $B^{(d)}$ bytes, and $B = \sum_{d=1}^{D} B^{(d)}$ is the total number of bytes.
Tokenizer $\tau$ cuts document $d$ into $N^{(d)}_\tau$ tokens $t^{(d)}_1, \dots, t^{(d)}_{N^{(d)}_\tau}$, each drawn from a vocabulary of $V_\tau$ possible tokens, with $N_\tau = \sum_{d=1}^{D} N^{(d)}_\tau$ as the total number of tokens.
Two tokenizers applied to the same corpus therefore agree on $B$ but differ in $N_\tau$ and $V_\tau$: the byte count belongs to the text, while the token count and the vocabulary size belong to the tokenizer.
A tokenizer-model pair $(\tau, p)$ obtains a total log loss of:
\begin{equation}\label{eq:total-loss}
    L(\tau, p) \;=\; \sum_{d=1}^{D} \; \sum_{n=1}^{N^{(d)}_\tau} -\log p\big(t^{(d)}_n \mid t^{(d)}_{<n}\big).
\end{equation}
One ubiquitous metric, the token-averaged negative log likelihood, divides by the token count $N_\tau$:
\begin{equation}\label{eq:nll}
    \mathrm{NLL}(\tau, p) \;=\; \frac{L(\tau, p)}{N_\tau}.
\end{equation}
The community has ruled out using $\mathrm{NLL}$ to compare models with different tokenizers because $\mathrm{NLL}$ divides the total loss by the number of tokens $N_\tau$, and different tokenizers cut the same text into different numbers of tokens.
Consider untrained models that, under $\mu$P and its modern successors, are approximately uniform over $V_\tau$ tokens  (\Cref{app:random-init}). For the uniform model $p^{\mathrm{unif}}$:
\begin{equation}\label{eq:nll-uniform}
    \mathrm{NLL}(\tau, p^{\mathrm{unif}}) \;=\; \frac{1}{N_\tau} \sum_{d=1}^{D} \; \sum_{n=1}^{N^{(d)}_\tau} -\log \frac{1}{V_\tau} \;=\; \log V_\tau.
\end{equation}

In this case, $\nll$ declares that the ``best'' model is the one with the smallest vocabulary size, even though all models are equally untrained.
\textbf{Token-averaged NLL is not agnostic to the tokenizer.}
As a solution, the research community turned to Bits-Per-Byte ($\bpb$)\footnote{For our purposes, the base (nats or bits) is irrelevant; changing base multiplies every score by the same positive constant and reorders nothing. Our maths present nats per byte for notational simplicity, and our experiments report $\bpb$ to match the literature and for numerical results to be correct.}, which divides the total log loss by the byte count $B$ rather than the token count $N_\tau$:
\begin{equation}\label{eq:bpb}
\bpb(\tau, p) \;=\; \frac{L(\tau, p)}{B}
\;=\;
\underbrace{\frac{N_\tau}{B}}_{\text{tokens per byte}}
\;\times\;
\underbrace{\frac{L(\tau, p)}{N_\tau}\rlap{\,.}}_{\text{token-averaged NLL}}
\end{equation}
Dividing by bytes removes the token count, but BPB fails the same test of untrained models.
To see why, let us again consider untrained models with different vocabulary sizes that place uniform probability on each of their $V_{\tau}$ tokens and ask whether $\bpb$ scores all model-tokenizer pairs equally:
\begin{equation}\label{eq:bpb-uniform}
\bpb(\tau, p^{\mathrm{unif}})
\;=\; \frac{L(\tau, p^{\mathrm{unif}})}{B}
\;=\; \frac{1}{B} \sum_{d=1}^{D} \; \sum_{n=1}^{N^{(d)}_\tau} -\log \frac{1}{V_\tau}
\;=\;
\underbrace{\frac{N_\tau}{B}}_{\text{tokens per byte}}
\;\times\;
\underbrace{\vphantom{\frac{N_\tau}{B}}\log V_\tau\rlap{\,.}}_{\text{nats per token}}
\end{equation}
$\bpb$ depends on the tokenizer in \emph{two} ways: (1) how many tokens the tokenizer produces per byte, and (2) how many bits the tokenizer-model pair spends per token.
The first factor decreases with vocabulary size while the second increases with it, and the two generally don't cancel: on a real English pretraining corpus \citep{penedo2024finewebdatasetsdecantingweb} with BPE tokenizers \citep{gage1994new, sennrich2016neural}, $\bpb$ is minimized near vocabulary size ${\sim}24$K (\Cref{fig:headline}); among untrained models that have learned nothing, $\bpb$ ranks some tokenizer-model pairs strictly above others.
Untrained models do not tie under $\bpb$, so $\bpb$ is not agnostic to the tokenizer either.

\begin{figure}[t!]
\centering
\includegraphics[width=\textwidth]{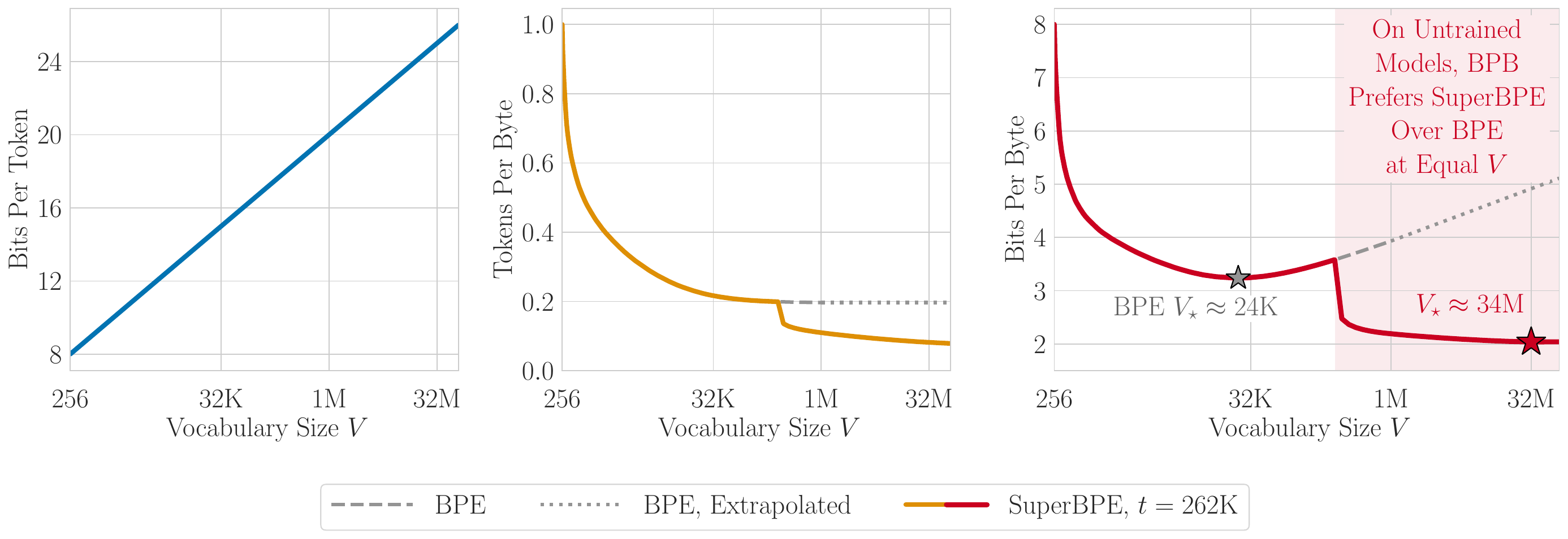}
\caption{\textbf{Myth \#1, Continued: Bits-Per-Byte Is Agnostic to the Tokenization Algorithm.}
The three panels of \Cref{fig:headline} for SuperBPE \citep{liu2025superbpe} (color) against standard BPE (gray, dashed), for untrained models on the same held-out corpus. Up to the transition point $t = 262$K the two tokenizers coincide. Above $t$, SuperBPE merges across pre-token boundaries and scores lower at every vocabulary size (shaded). Beyond $1.05$ million, where the BPE sweep ends, the dotted BPE curve holds tokens per byte fixed. BPB prefers one tokenization algorithm (SuperBPE) over another (BPE), even for untrained models.
Corpus: FineWeb-Edu-dedup, English \citep{penedo2024finewebdatasetsdecantingweb}.
}
\label{fig:superbpe}
\end{figure}

If a metric were agnostic to the tokenizer, models that have learned nothing would tie.
They do not.
On this corpus the minimum sits near $24$K.
On four other corpora (code, French, Hindi, and Russian), with the same BPE pipeline trained on each, the minimum sits anywhere from 504 (Hindi) to $117$K (Russian; \Cref{tab:corpus-minima} in \Cref{app:sweep-setup}).
On a fifth, Chinese, no minimum appears in a sweep to 1.05 million.
Under SuperBPE \citep{liu2025superbpe}, which merges across pre-token boundaries once the vocabulary passes a transition point $t$, the minimum moves to about $34$ million for $t = 262$K (\Cref{fig:superbpe}), a stress test of the metric rather than a recommendation.
From $t$ to $1.05$ million, where our BPE sweep ends, SuperBPE scores lower than BPE at every vocabulary size.
At $1.05$ million, the gap between the untrained models is $1.74$ $\bpb$.
At a vocabulary size of $200$K, the released BPE and SuperBPE ($t = 180$K) tokenizers of \citet{liu2025superbpe} differ by $0.88$ $\bpb$ on the same English corpus, again for untrained models.
What matters is the preference, not its location: when the model has learned nothing, the metric still prefers some tokenizer families and vocabulary sizes over others, and that preference is tokenizer dependence.

\section{Myth \#2: Lower Bits-Per-Byte Means a Better Model}
\label{sec:downstream}

Lower $\bpb$ means a shorter code for the evaluated text; it does not imply better downstream performance.
We show this with standard evaluation protocols.

\begin{figure}[t!]
\centering
\includegraphics[width=\textwidth]{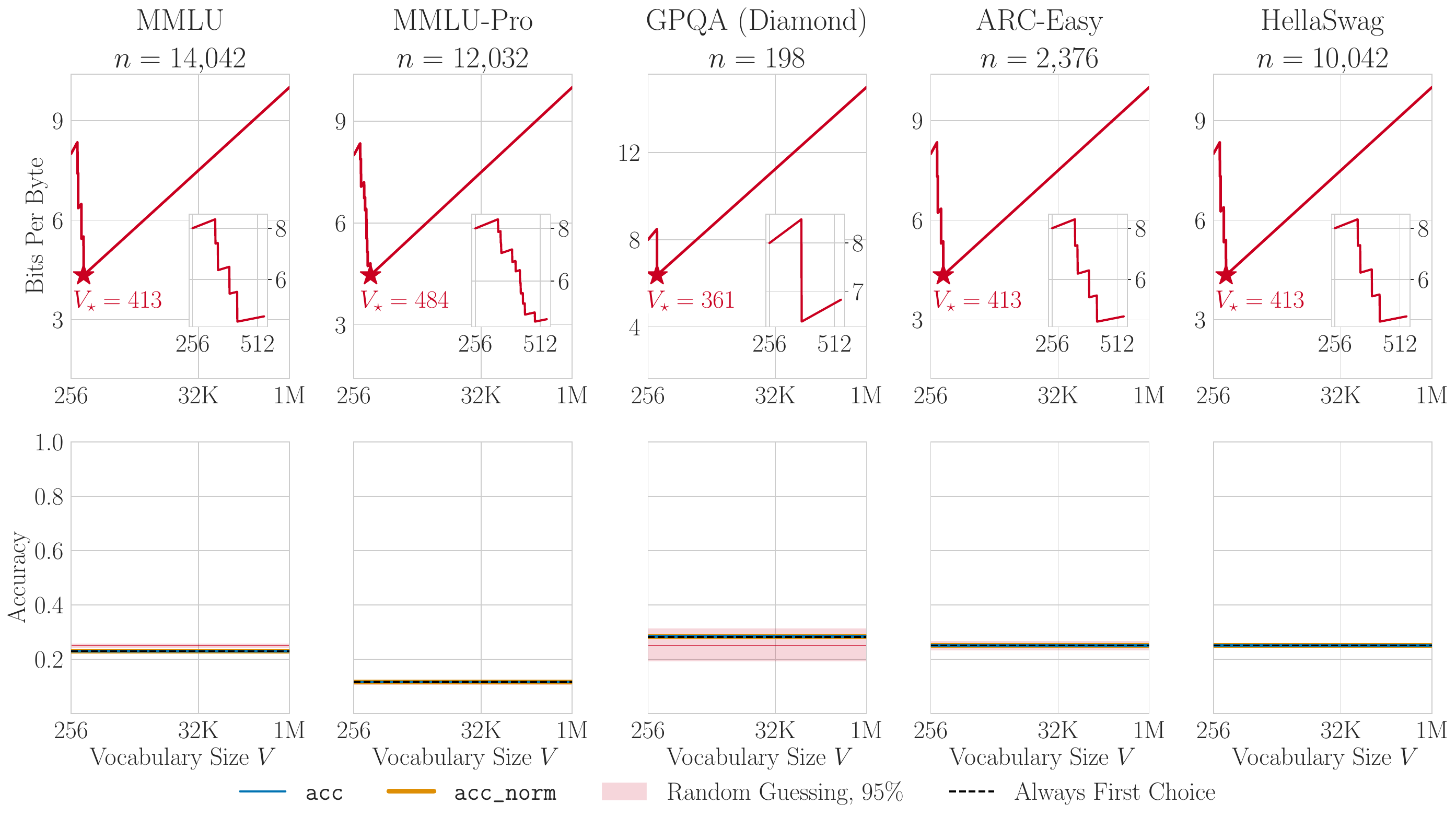}
\caption{\textbf{Myth \#2: Lower Bits-Per-Byte Means a Better Model (Option-Letter Scoring).}
Untrained models that output uniform distributions.
Top: $\bpb$ of the correct label drops one step at each merge that shortens a label (inset), then rises with $\log V$.
Bottom: accuracy is constant, at the fraction of items whose correct answer is listed first.
Red band, for reference: central $95\%$ of a uniform random guesser; black dashed: always choosing the first option.
$\bpb$ strongly prefers a vocabulary of a few hundred tokens, whereas accuracy finds all models to be equally useless.}
\label{fig:mcqa-letter}
\end{figure}

We construct $81$ nested BPE tokenizers, with vocabulary sizes ranging from $2^8$ to $2^{20}$, and score uniform distributions on standard tasks: MMLU \citep{hendrycks2021mmlu}, MMLU-Pro \citep{wang2024mmlupro}, GPQA \citep{rein2024gpqa}, ARC \citep{clark2018arc}, HellaSwag \citep{zellers2019hellaswag}, PIQA \citep{bisk2020piqa}, and WinoGrande \citep{sakaguchi2020winogrande}.

\begin{table}[b!]
\caption{\textbf{Two Standard Ways to Score Multiple-Choice Question-Answering Problems.}
Example problem from ARC-Easy \citep{clark2018arc}.
The correct option is bold, and \textvisiblespace\ marks a leading space.
Accuracy compares the four scored strings; $\bpb$ charges the correct one.}
\label{tab:mcqa-formats}
\centering
\small
\begin{tabular}{@{}p{9em}>{\raggedright\arraybackslash}p{15em}>{\raggedright\arraybackslash}p{15em}@{}}
\toprule
& \textbf{Option-Letter Scoring} & \textbf{Answer-Text Scoring} \\
\midrule
Prompt
& \texttt{Question: Which piece of safety equipment is used to keep mold spores from entering the respiratory system?}\newline
  \texttt{A. safety goggles}\newline
  \texttt{B. breathing mask}\newline
  \texttt{C. rubber gloves}\newline
  \texttt{D. lead apron}\newline
  \texttt{Answer:}
& \texttt{Question: Which piece of safety equipment is used to keep mold spores from entering the respiratory system?}\newline
  \texttt{Answer:} \\
\addlinespace
Scored strings
& \texttt{\textvisiblespace A}, \ \textbf{\texttt{\textvisiblespace B}}, \ \texttt{\textvisiblespace C}, \ \texttt{\textvisiblespace D}
& \texttt{\textvisiblespace safety goggles}, \ \textbf{\texttt{\textvisiblespace breathing mask}}, \ \texttt{\textvisiblespace rubber gloves}, \ \texttt{\textvisiblespace lead apron} \\
\addlinespace
Bytes charged
& $2$ & $15$ \\
\bottomrule
\end{tabular}
\end{table}

On multiple-choice question-answering benchmarks, there are at least two standard ways to score models:
(1) In \emph{option-letter scoring}, the prompt lists the options under labels, and the model is scored on a label;
(2) in \emph{answer-text scoring}, the prompt shows only the question, and the model is scored on the text of each answer.
\Cref{tab:mcqa-formats} shows both.
In both, lm-evaluation-harness appends each option to the prompt and computes its log-likelihood \citep{eval-harness};
\texttt{acc} picks the option with the highest log-likelihood and \texttt{acc\_norm} the highest per character, with ties going to the first.
$\bpb$ reads the same strings: it charges the code length of each correct option, given the prompt, and divides by the bytes of those options.
Accuracy and $\bpb$ therefore differ only in the metric.

\begin{figure}[t!]
\centering
\includegraphics[width=\textwidth]{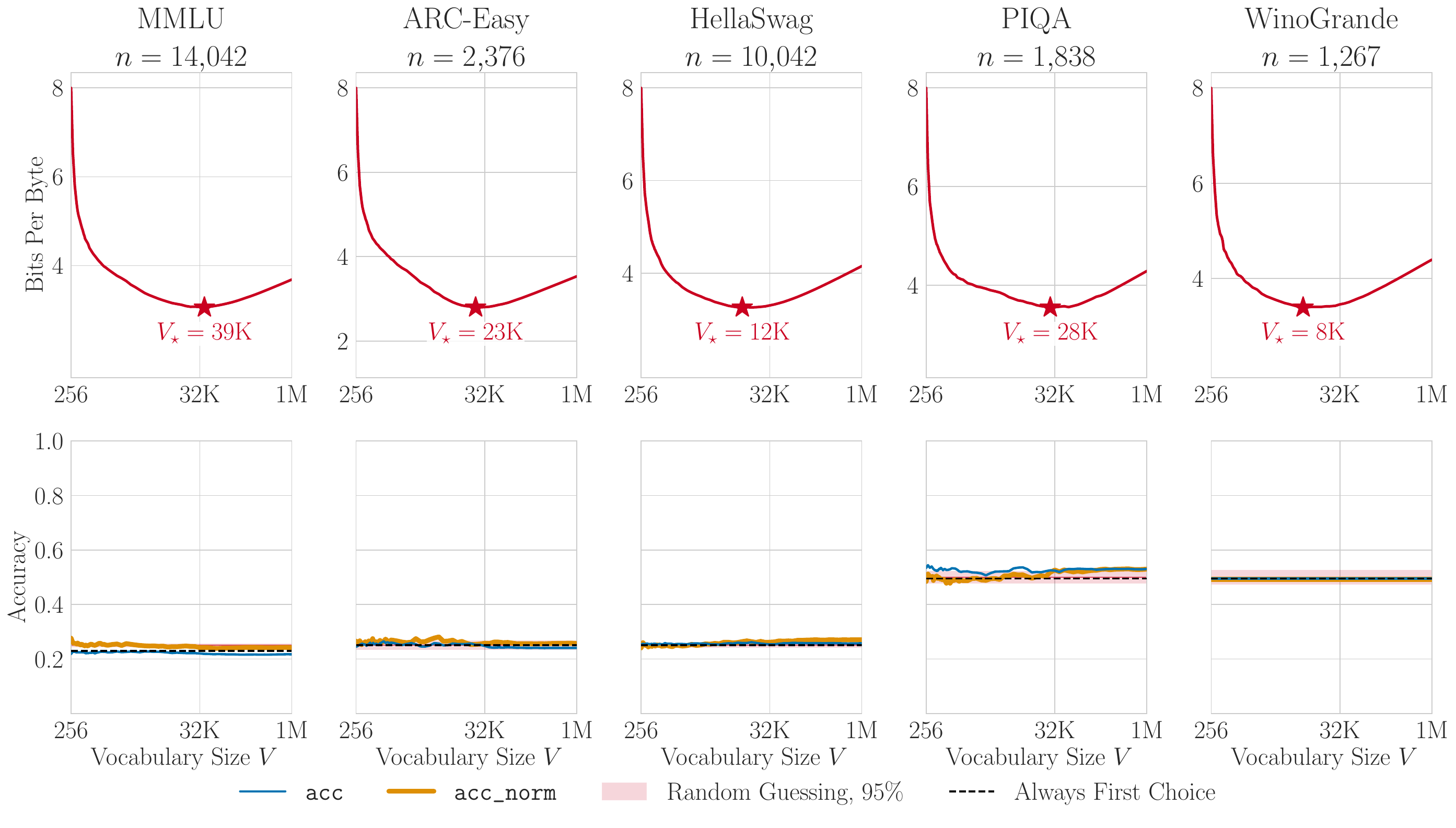}
\caption{\textbf{Myth \#2: Lower Bits-Per-Byte Means a Better Model (Answer-Text Scoring).}
Untrained models that output uniform distributions, scored on answer text (WinoGrande: on a shared continuation).
Top: $\bpb$ of the correct answers has an interior minimum on every benchmark (star), between $8$K and $39$K.
Bottom: accuracy on the same strings moves by at most a few points and has no common preferred vocabulary size.
Red band, for reference: central $95\%$ of a uniform random guesser; dashed: always choosing the first option.
$\bpb$ concludes that vocabulary sizes between $8$K and $39$K are strongly preferable, whereas accuracy finds all models to be equally useless.
}
\label{fig:mcqa-continuation}
\end{figure}

\textbf{In both approaches, code length strongly prefers a vocabulary size whereas accuracy does not.}
On option labels, accuracy is exactly constant (\Cref{fig:mcqa-letter}): the first label is never longer in tokens than the others, so the uniform model always picks it; MMLU accuracy is $0.2295$, the fraction of items whose answer is listed first, at every vocabulary size, while its code length moves between $4.34$ and $10.00$ bits-per-byte.
On answer text, $\bpb$ is minimized between $8$K and $39$K (\Cref{fig:mcqa-continuation}), whereas accuracy is flat at chance and never peaks at the $\bpb$ minimum.
The reason is that the uniform model charges every token the same $\log V_\tau$: when accuracy compares the options of one item, $\log V_\tau$ cancels, but $\bpb$ multiplies it by tokens per byte.
To understand where the jagged edges of \Cref{fig:mcqa-letter} emerge from, see a simple mathematical derivation in \Cref{app:downstream_derivation}.

These measurements concern models that have learned nothing.

\begin{figure}[t]
\centering
\includegraphics[width=\textwidth]{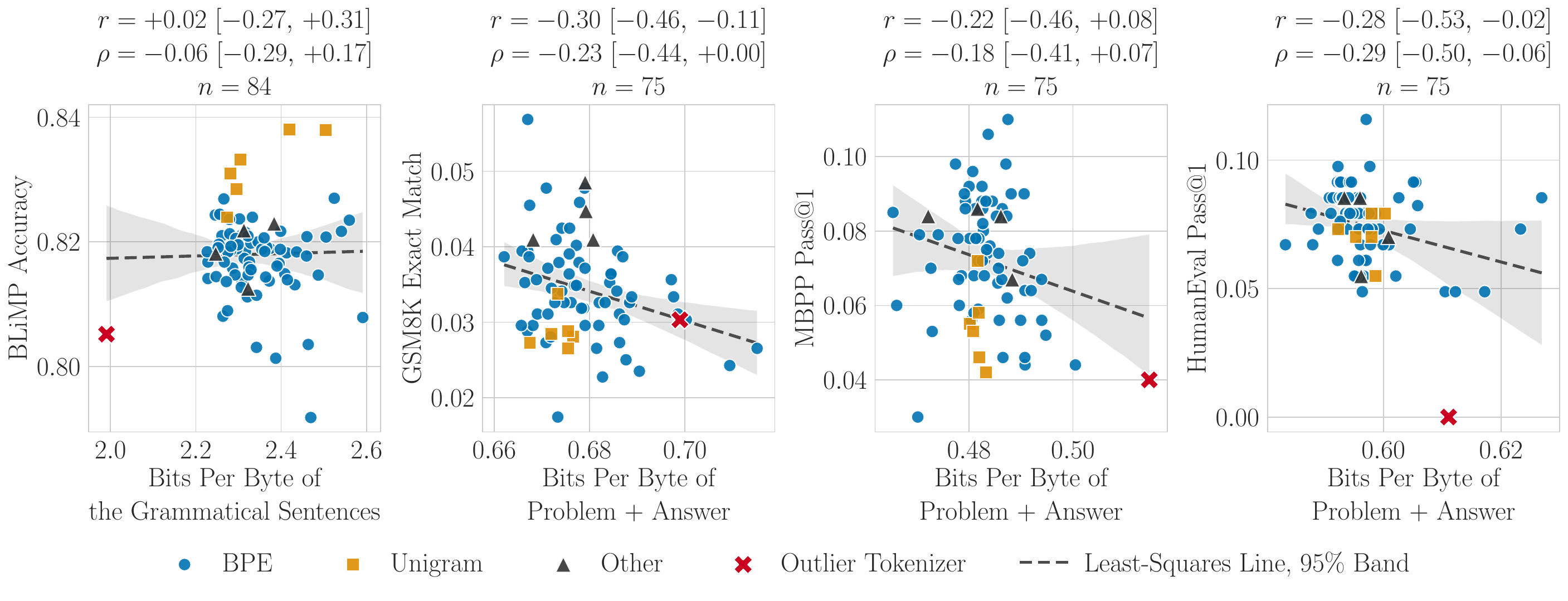}
\caption{\textbf{Myth \#2: Lower Bits-Per-Byte Means a Better Model (Trained Models).}
Each point is one model-tokenizer pair from \citet{meister2026tokeval}: every model has the same layers and saw the same documents in the same order; the tokenizer varies.
The horizontal axis is $\bpb$ on the benchmark's own text; for BLiMP, the grammatical sentence of each pair, scored alone.
Each panel gives the Pearson ($r$) and Spearman ($\rho$) correlations between $\bpb$ and downstream performance.
Dashed line and band: least-squares line with a $95\%$ bootstrap band.
On trained models, the correlation is relatively weak, with or without the outlier tokenizer (red cross, \Cref{app:tokeval_grid}).
}
\label{fig:tokeval-grid}
\end{figure}

\textbf{Among trained models that differ in the tokenizer, $\bpb$ and task scores are weakly coupled on the same text.}
We use the model-tokenizer pairs of \citet{meister2026tokeval}: one 24-layer network trained with 85 tokenizers\footnote{One tokenizer is an outlier; without it, no Pearson correlation moves by more than $0.093$ (\Cref{app:tokeval_grid}).} on the same documents in the same order; the tokenizer also changes the parameter count (0.98B to 1.60B) and the number of tokens trained on (9.0B to 10.5B).
On four standard benchmarks, BLiMP \citep{warstadt2020blimp}, GSM8K \citep{cobbe2021gsm8k}, MBPP \citep{austin2021mbpp}, and HumanEval \citep{chen2021humaneval}, we measure each model's $\bpb$ on each benchmark's own text: every problem followed by its reference answer, in the prompt format that produced the released scores, and for BLiMP the grammatical sentence of each pair.
We score the reference answer because the log-likelihood of gold reference solutions tracks pass@$k$ on generative benchmarks as pretraining scales \citep{schaeffer2026generative}.
$\bpb$ and the score are then computed from the same benchmark.
See \Cref{app:tokeval_grid} for more information.

The correlations are weak and noisy on every benchmark (\Cref{fig:tokeval-grid}).
The Pearson correlation between $\bpb$ and the score is $+0.02$ for BLiMP, with a $95\%$ bootstrap interval of $[-0.27, +0.31]$; $-0.22$ for MBPP; $-0.28$ for HumanEval; and $-0.30$ for GSM8K; the intervals for BLiMP and MBPP include zero.
Even the strongest, on GSM8K, is weak: it accounts for less than a tenth of the variance in the score.
This holds for a grammaticality benchmark scored by log-likelihood, a math benchmark scored by exact match, and two code benchmarks scored by pass@1.
While the released scores are noisy\footnote{Across tokenizers, a score spreads only $1.2$ to $2.4$ times its run-to-run noise (\Cref{app:tokeval_grid}).}, BLiMP, the least noisy score, is also the least correlated.

\textbf{$\bpb$ charges for every bit of the correct answer, but a task score depends on only part of those bits.}
Five published tokenizer sweeps and one comparison of byte-level with token-level models report that $\bpb$ and task scores disagree, and two of them suggest a mechanism (\Cref{app:related_work}).
To see where the bits go, we split the code length of the correct answer into the part that a task score reads and the part that it ignores.
On a generative benchmark, the split cannot be computed exactly, because many strings reach the same outcome and their total probability cannot be summed.
On a multiple-choice benchmark, it can, because the options are the only strings that are scored.
For an item with prompt $x$, correct string $y^\star$ and correct option $c^\star$, let $F$ be the event that the response is one of the options and $Y$ the option it selects.
The chain rule splits the code length of the correct string into three terms:
\begin{equation}\label{eq:answer-factorization}
\underbrace{-\log p(y^\star \mid x)}_{\text{what } \bpb \text{ charges}}
\;=\;
\underbrace{-\log p(F \mid x)}_{\text{format compliance}}
\;+\;
\underbrace{-\log p(Y {=} c^\star \mid F, x)}_{\text{answer discrimination}}
\;+\;
\underbrace{-\log p(y^\star \mid Y {=} c^\star, F, x)}_{\text{exact rendering}} .
\end{equation}
We call these the format term, the decision term, and the rendering term.
Under forced choice, accuracy (\texttt{acc}) depends only on the distribution over the options, which the format term does not change.
For the uniform model of \Cref{fig:mcqa-continuation}, MMLU's correct answers average $44.3$ bytes and cost about $135$ bits at the vocabulary size that minimizes its $\bpb$, about $39$K: $95.5$ in the format term and $40.0$ in the decision term, because this model favors the option with the fewest tokens.
Moving the vocabulary from $39$K to $2^{20}$ adds $27.8$ bits to the same answers, $20.0$ of them in the format term, while accuracy stays at chance.

\begin{figure}[t]
\centering
\includegraphics[width=\textwidth]{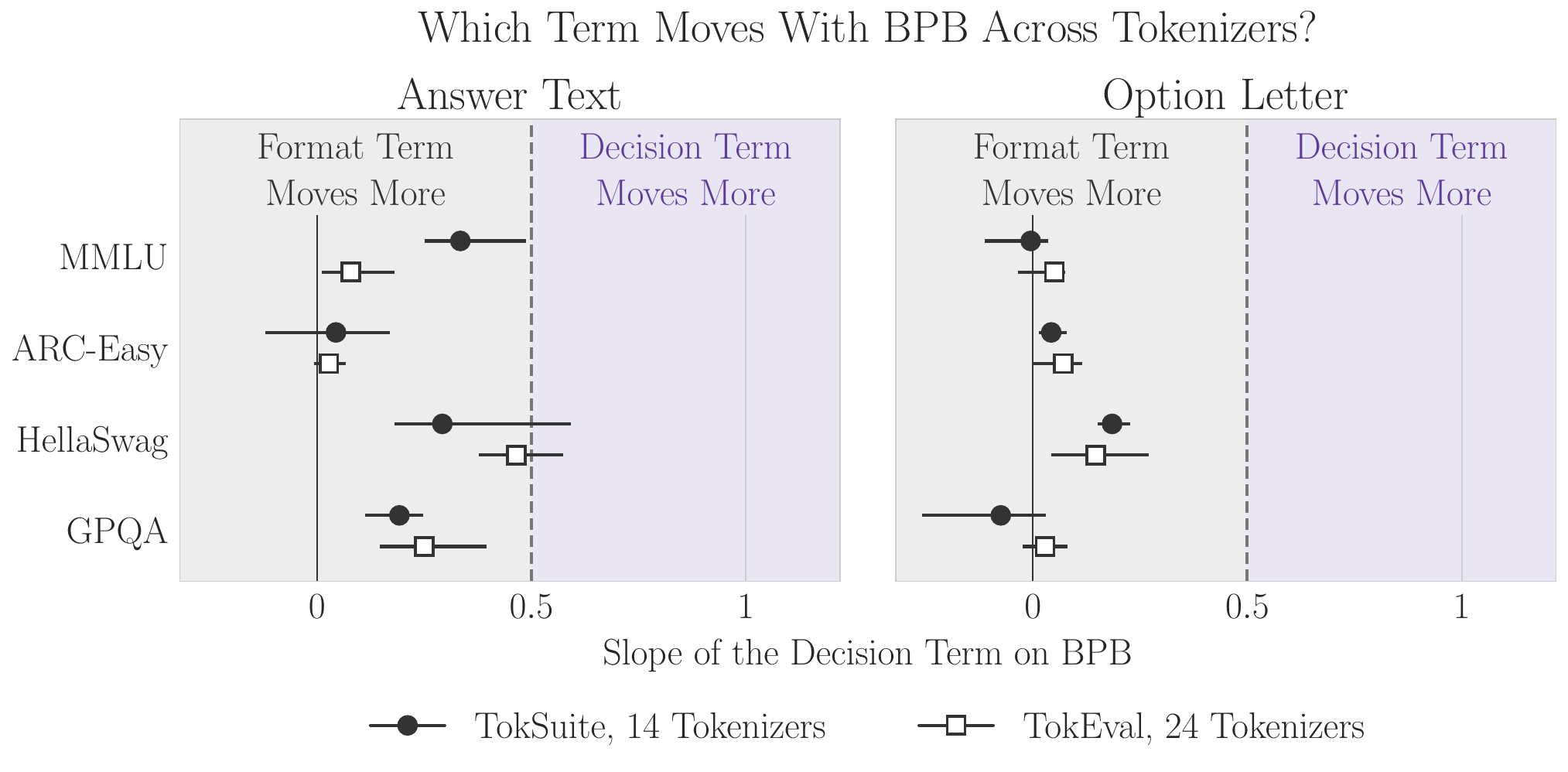}
\caption{\textbf{Myth \#2, Continued: Lower Bits-Per-Byte Means a Better Model (Where the Differences Lie).}
Each marker is the least-squares slope of the decision term on $\bpb$ across the tokenizers of one set, with its $95\%$ bootstrap interval; the format term's slope is one minus it.
Most of the difference lies in the format term, which accuracy ignores, not in the decision term, which it reads.}
\label{fig:decision-share}
\end{figure}

\textbf{On trained models, most of the $\bpb$ difference between tokenizers lies in the format term.}
If a change of tokenizer moved only the format term, $\bpb$ would move and accuracy would not.
We measure which term moves more on 49 released models from two sets, in each of which the tokenizer varies: TokSuite \citep{altintas2026toksuite}, 16 runs of 14 tokenizers trained for the same number of tokens, and TokEval \citep{meister2026tokeval}, 33 runs of 24 tokenizers trained on the same documents in the same order.
Each model is scored on four benchmarks, MMLU, ARC-Easy, HellaSwag, and GPQA, once on the option letter and once on the answer text.
Each option is one string, so the rendering term is zero, and the other two terms can be computed exactly (\Cref{app:decision_format}).
\Cref{fig:decision-share} gives the slope of the decision term on $\bpb$ across tokenizers in all sixteen conditions.

In all sixteen conditions, the decision term's slope is below one half, so the format term carries more of the difference; only on HellaSwag answer text does the slope's interval reach above one half.
On answer text, the decision term's slope lies between $0.03$ and $0.46$.
On MMLU letters, the decision term is the larger term but carries almost none of the difference: it is $83\%$ of the $\bpb$ in TokSuite and $75\%$ in TokEval, yet its slope is $0.05$ or less in both.
Every model is at chance on option letters, so the decision term stays near its chance value whatever the tokenizer.

\paragraph{A strictly proper score on the exact text must charge for rendering.}
A task sees only the outcome a response reaches; $\bpb$ sees the exact string.
Fix a prompt, and let a model give the distribution $Q$ over exact responses; in \Cref{eq:answer-factorization}, $Q = p(\cdot \mid x)$.
Let $o(y)$ be the outcome of response $y$, such as the option it selects or whether a program passes its tests.
The model then gives each outcome $a$ the probability $\bar{Q}(a)$, the sum of $Q(y)$ over the responses $y$ with $o(y) = a$.
A score that measures only the task depends on $Q$ only through $\bar{Q}$.
A score is \emph{strictly proper} if, for every true distribution $P$, the prediction $P$ gets a strictly better expected score than every other prediction \citep{gneiting2007proper}; the log loss behind $\bpb$ is strictly proper.

\begin{proposition}[No score is both strictly proper and blind to rendering]
\label{prop:outcome-sufficiency}
If two or more responses have the same outcome, no score is both strictly proper for the distribution over exact responses and a function of $\bar{Q}$ alone.
\end{proposition}

\begin{proof}
Choose an outcome $a$ that two or more responses reach.
Let $Q_1 \neq Q_2$ give every outcome the same probability but spread the probability of $a$ differently over the responses that reach it.
Then $\bar{Q}_1 = \bar{Q}_2$, so a function of $\bar{Q}$ gives $Q_1$ and $Q_2$ the same expected score under every true distribution.
But under $P = Q_1$, a strictly proper score gives $Q_1$ a strictly better expected score than $Q_2$.
\end{proof}

The two predictions in the proof differ only in the rendering term of \Cref{eq:answer-factorization}.
This term matters when many strings reach one outcome, as in the GSM8K, MBPP, and HumanEval scores of \Cref{fig:tokeval-grid}: many solutions reach the same final number, and many programs pass the same tests.
There, $\bpb$ on the reference answer charges for how the model spreads probability over those strings, and the score does not, so the two can move apart.
A score on outcomes can still be strictly proper: on a multiple-choice item, the decision term is the log loss of the distribution over options, which is the distribution that accuracy reads.
Accuracy itself reads only which option is largest, so it is not strictly proper even over options.

\section{Myth \#3: Different \texorpdfstring{``Tokenizer-Agnostic''}{"Tokenizer-Agnostic"} Metrics Yield Consistent Results}
\label{sec:unigram_normalized_loss}

\begin{figure}[t]
\centering
\includegraphics[width=\textwidth]{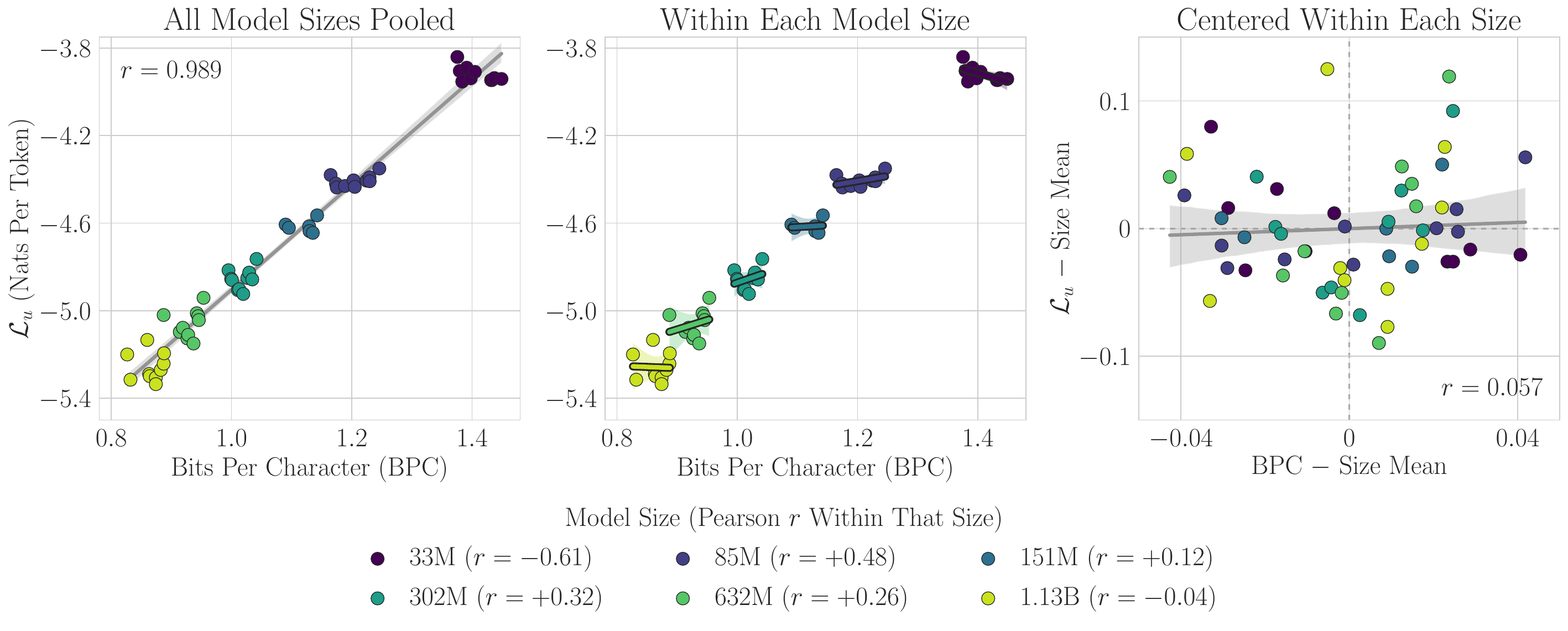}
\caption{\textbf{Myth \#3: Different ``Tokenizer-Agnostic'' Metrics Yield Consistent Results (Correlation).}
The $56$ points of Figure~11 of \citet{tao2024scalingvocab}, digitized and colored by model size, with least-squares fits and $95\%$ confidence bands (bootstrap on the left and right; analytic in the center, where each model size has few points).
Left: pooled, $\mathcal{L}_u$ and BPC correlate at $r = 0.989$.
Center: within each model size, where the points differ in vocabulary size, the correlation ranges from $-0.61$ to $+0.48$.
Right: with each size's mean removed from both metrics, $r = 0.057$. The pooled correlation comes from differences between model sizes; within a size, little remains.}
\label{fig:unigram-simpson}
\end{figure}

Not all tokenizer papers use Bits-Per-Byte.
\citet{roh2020unigramnorm} proposed the unigram-normalized perplexity to compare language models with different vocabulary sizes, and \citet{tao2024scalingvocab} reformulated it as a loss to fit scaling laws for the vocabulary size:
\begin{equation}\label{eq:unigram-normalized-loss}
    \mathcal{L}_u(\tau, p) \;=\; \frac{1}{N_\tau} \sum_{d=1}^{D} \; \sum_{n=1}^{N^{(d)}_\tau} -\log \frac{p\big(t^{(d)}_n \mid t^{(d)}_{<n}\big)}{u_\tau\big(t^{(d)}_n\big)},
\end{equation}
where $u_\tau(t)$ is the frequency of token $t$.
$\mathcal{L}_u$ is the model's loss per token minus that of the tokenizer's unigram distribution.
To justify the metric, \citet{tao2024scalingvocab} report that $\mathcal{L}_u$ is highly correlated with bits per character (BPC; on one text, BPC is $\bpb$ times a constant): Pearson $r = 0.989$ across their models (their Figure~11; reproduced in our \Cref{fig:unigram-simpson}, left).

This claim has three problems.

\paragraph{Unigram-normalized loss is not tokenizer-agnostic.}

\begin{figure}[t]
\centering
\includegraphics[width=\textwidth]{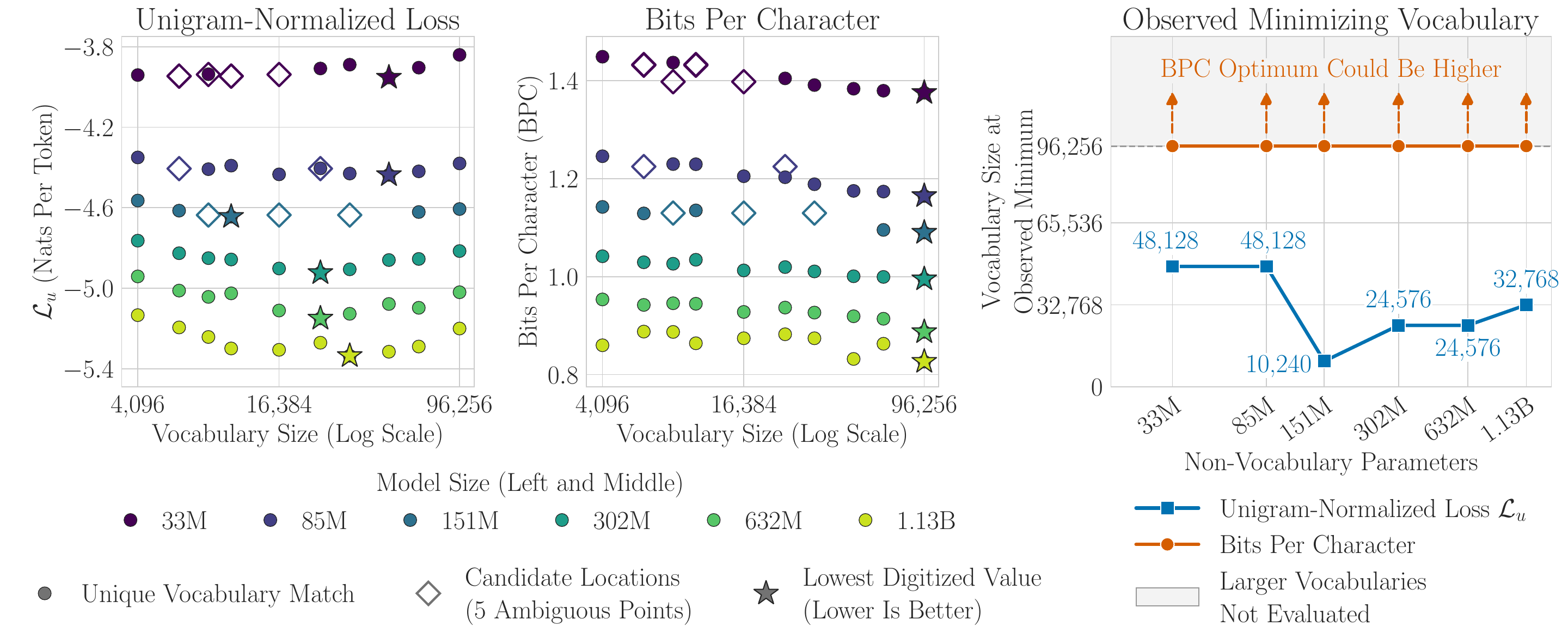}
\caption{\textbf{Myth \#3: Different ``Tokenizer-Agnostic'' Metrics Yield Consistent Results (Minima).}
Left and center: the $56$ digitized runs of Figure~11 of \citet{tao2024scalingvocab} against vocabulary size; stars mark each model size's minimum, and hollow diamonds show candidate vocabularies for five runs whose match is ambiguous.
Right: the vocabulary at each minimum.
BPC is always lowest at the largest vocabulary trained, and its optimum may lie beyond it (arrows); $\mathcal{L}_u$ is lowest inside the grid.
Unigram-normalized loss and BPC pick different vocabularies from the same data.}
\label{fig:tao-vocab-minima}
\end{figure}

\citet{roh2020unigramnorm} call their metric ``independent of the vocabulary size'' and conclude that a difference in it ``purely comes from the ability of LM''.
\citet{tao2024scalingvocab} call their loss ``vocabulary-insensitive'' and write that the difference in $\mathcal{L}_u$ ``comes from the ability of the language model itself''.
A 2026 survey of tokenization lists $\mathcal{L}_u$ among the metrics that ``can be used to compare across different tokenizers'' and writes that normalizing the loss ``by the vocabulary's unigram distribution [...] corrects for vocabulary size'' \citep{cognetta2026tokenizationsurvey}.
However, by the same argument as for $\bpb$, none of these claims hold.
Splitting the logarithm in \Cref{eq:unigram-normalized-loss} gives
\begin{equation}\label{eq:unigram-normalized-loss-bpb}
    \mathcal{L}_u(\tau, p)
    \;=\; \underbrace{\frac{B}{N_\tau}}_{\text{bytes per token}}
    \times
    \Big[\bpb(\tau, p) - \bpb(\tau, u_\tau)\Big].
\end{equation}
$\mathcal{L}_u$ is the difference between the model's $\bpb$ and that of its tokenizer's unigram model, converted back to a per-token quantity.
It measures improvement over a reference, a legitimate question; what fails is the claim that it removes tokenizer effects.
Changing the tokenizer changes all three terms: the model's BPB, the unigram reference, and the bytes per token, the same per-token denominator that the community rejected for $\mathrm{NLL}$.

\paragraph{The correlation between metrics comes from model size, not from the metrics.}
When computing the correlation between unigram-normalized loss and BPB, \citet{tao2024scalingvocab} use multiple model sizes from $33$M to $1.13$B (non-vocabulary) parameters.
As scaling laws tell us, larger models reach lower losses \citep{kaplan2020scaling, hoffmann2022training}.
This produces a strong global correlation between $\mathcal{L}_u$ and $\bpb$ (\Cref{fig:unigram-simpson}, left).
We digitized Figure~11 and labeled each point with its model size.\footnote{Coordinates are accurate to one pixel. Each point matches the released final-checkpoint losses of exactly one model size (\texttt{sail-sg/scaling-with-vocab}, commit \texttt{6701cc2}).}
However, the point of $\mathcal{L}_u$ is to compare vocabularies \emph{at a fixed model size}, which pooling model sizes cannot test.
Within each model size, the correlation ranges from $-0.61$ to $+0.48$ (\Cref{fig:unigram-simpson}, center).
With each size's mean removed, the Pearson correlation falls to $0.057$ (\Cref{fig:unigram-simpson}, right).\footnote{Random shifts of every digitized point by up to one pixel keep the centered $r$ between $0.046$ and $0.067$ and change no minimum of \Cref{fig:tao-vocab-minima}.}

\paragraph{The two metrics select different vocabularies from the same data.}

Even if neither metric is tokenizer-agnostic, one might hope that they select the same vocabularies.
However, this is not the case.
$\mathcal{L}_u$ is lowest inside the grid, between $10{,}240$ and $48{,}128$, and its minimum moves up and down with model size (\Cref{fig:tao-vocab-minima}, left).
In comparison, BPC is lowest at the largest vocabulary trained for every model size, $96{,}256$: the best evaluated point, not a located optimum (\Cref{fig:tao-vocab-minima}, center).
Thus, from the same data, the two metrics select different vocabularies at each model size.
\citet{tao2024scalingvocab} also test the vocabulary that their law predicts on downstream tasks (their Table~2), but this test does not settle the disagreement: it compares one trained model per vocabulary, their own law says the two models have almost the same loss, and the predicted vocabulary comes from a units error in the released fitting code (\Cref{app:tao_table2}).

\section{Can Any Metric Be Tokenizer-Agnostic? The Tokenizer Trilemma}
\label{sec:impossibility}

All three myths come from metrics that fail in unexpected ways.
One might then ask whether any metric has three obviously desirable properties.
None does.

Every tokenizer-model pair $(\tau, p)$ assigns a probability to every text; call these its \emph{predictions}.
We would like a metric to have three properties:
\begin{itemize}[leftmargin=2.6em, itemsep=1pt, topsep=2pt]
    \item[\textbf{(P1)}] \textbf{Only the truth scores best}: with the tokenizer fixed, every prediction other than the true data distribution gets a worse expected score;
    \item[\textbf{(P2)}] \textbf{Same predictions, same score}: two pairs that give every text the same probability get the same score, however their tokenizers split the text;
    \item[\textbf{(P3)}] \textbf{Untrained models tie}: models that have learned nothing, such as $p^{\mathrm{unif}}$, get the same score under every tokenizer.
\end{itemize}
$\mathrm{NLL}$ breaks (P2), because two pairs with the same predictions can split a text into different numbers of tokens, and it breaks (P3) by \Cref{eq:nll-uniform}.
$\bpb$ repairs (P2), because its denominator $B$ belongs to the text, not to the tokenizer.
This is why the community switched to $\bpb$, and why it believes $\bpb$ is tokenizer-agnostic.
But \Cref{fig:headline} shows that $\bpb$ breaks (P3).

\textbf{Why no metric can have all three.}
Uniform pairs do not make the same predictions.
BPE builds its vocabulary from data, so common byte sequences become single tokens, and a uniform model over that vocabulary gives common text a higher probability before any training.
\textbf{The model has learned nothing, but the pair has.}
A metric with (P1) and (P2) judges pairs only by their predictions, and it must tell different predictions apart.
So it cannot score these uniform pairs as equal: one tokenizer gets a head start.
A tokenizer that learns nothing from data shows the contrast: cut text into blocks of $k$ bytes, and every uniform pair scores exactly $8$ bits per byte, for every $k$.\footnote{By \Cref{eq:bpb-uniform}: $V_\tau = 256^k$ and $N_\tau/B = 1/k$, exactly when $k$ divides every document's length.}

\begin{theorem}[The tokenizer trilemma]
\label{thm:trilemma}
If two lossless tokenizers, each with its untrained reference model, make different predictions, no metric has all of (P1), (P2), and (P3).
\end{theorem}

\begin{proof}[Proof sketch]
Let text be drawn from the predictions of the first uniform pair.
Because the first tokenizer is lossless, some autoregressive model on it, not necessarily of a given size, copies the predictions of the second uniform pair.
By (P1), the first uniform pair beats that copy; by (P2), the copy scores the same as the second uniform pair.
So the two uniform pairs do not tie, which breaks (P3).
\Cref{app:trilemma} gives the full result, in which each untrained reference is conditioned on the length of the text.
\end{proof}

\textbf{Every metric in this paper gives up at least one property, and each choice answers a different question.}
$\bpb$ keeps (P1) and (P2) and gives up (P3), which is Myth~\#1; it answers \emph{which pair encodes this text in fewer bits?}
Subtracting each tokenizer's untrained score keeps (P1) and (P3) and gives up (P2); it answers \emph{how far did each model move from its own tokenizer's starting point?}
The unigram-normalized loss of Myth~\#3 is a score of this second kind, with each tokenizer's unigram model as the reference.
Metrics that give up different properties can rank the same runs differently, as Myth~\#3 shows.
Keeping (P2) and (P3) means giving some different predictions the same score, which gives up (P1).
Code length and accuracy face the same conflict at the level of the task (\Cref{prop:outcome-sufficiency}, Myth~\#2): a score under which only the truth scores best must tell every two different predictions apart, so a coarser question, such as which option a model picks, needs a score on coarser outcomes.
No metric has all three properties; picking a metric is picking a question.

\section{Discussion}
\label{sec:recommendations}

\paragraph{Read the Tokenization Literature with Caution.}
The tokenization literature needs careful reading: results that appear to conflict often answer different questions.
Modifying tokenization can introduce several degrees of freedom at once, for example by decoupling compute from the number of unique training bytes, and this makes experiments hard to compare.
Papers can rest a claim on weak evidence: \citet{tao2024scalingvocab} support their predicted vocabulary with a downstream gain between two vocabularies to which their own released law assigns almost the same loss (\Cref{app:tao_table2}).
Papers can also be mutually inconsistent: \citet{tao2024scalingvocab} and \citet{limisiewicz2026computeoptimal} reach opposing prescriptions for how to optimally scale vocabulary size, but without identifying or explaining the factors responsible for the apparent contradiction.

$\bpb$ measures one thing well: how many bits a tokenizer-model pair needs to encode a text.
Several of the papers we quote use it for exactly that, even where they call it tokenizer-agnostic.
The myths begin when that fact is read as the answer to different questions.

\paragraph{Intelligence as Discrimination, Not Compression.}

The machine learning research community has frequently viewed intelligence as prediction or, equivalently, compression.
This view draws on Solomonoff induction, Kolmogorov complexity, and the Minimum Description Length principle, which formalize Occam's razor as a preference for the shortest description \citep{solomonoff1964formal,kolmogorov1965three,rissanen1978modeling}.
It continues in theories of universal intelligence \citep{hutter2005universal} and in work that measures language models by how well they compress \citep{deletang2024language,huang2024compression}.

We conjecture that this view is correct only near the true data distribution, where the distribution with the shortest expected code is also the one that most often identifies the correct continuation.
Far from the true data distribution, intelligence is the ability to discriminate: to identify the correct continuation or the correct action from plausible incorrect continuations or plausible incorrect actions.
There, compression and discrimination come apart.
A model can shorten its code by spending fewer bits on format and rendering (\Cref{eq:answer-factorization}), which do not help it identify the correct continuation.
And the decision term is a log-probability of the correct choice, while identifying that choice requires only that every plausible incorrect choice receive less mass \citep{schaeffer2024elusive}.
The models of \Cref{sec:downstream} are far from the true data distribution, and in them the two have already come apart.
Across models that differ in the tokenizer, the code length moves mostly with the format term, and the decision term carries a minority of its spread (\Cref{fig:decision-share}).
So most of what separates these models in $\bpb$ is not their ability to discriminate.
If the conjecture holds, compression is a proxy for intelligence only near the true data distribution, and a measure of intelligence far from it must score discrimination directly.

\clearpage

\bibliography{references}
\bibliographystyle{iclr2027_conference}

\clearpage

\appendix
\crefalias{section}{appendix}
\crefalias{subsection}{subappendix}
\section{Related Work}
\label{app:related_work}

A language model assigns a probability to each unit of text in turn, and its tokenizer decides what those units are.
A tokenizer is a set of choices: the size of its vocabulary, the algorithm that builds the vocabulary from a corpus, the pre-tokenization rule that splits text into pieces such as words, numbers and punctuation before the algorithm runs, and the rule that picks one segmentation of each piece.
Studies vary each choice in turn: the algorithm \citep{bostrom2020byte}, the pre-tokenization rule \citep{dagan2024getting}, and the inference rule \citep{uzan2024greed}.
Together these choices set how many tokens a text becomes, often reported as fertility, the number of tokens per word, and how many alternatives each prediction chooses among, the vocabulary size $V$.
For a model that has learned nothing, these are the two factors of $\bpb$ in \Cref{eq:bpb-uniform}.
Because the vocabulary is learned from a corpus, a tokenizer carries information about text before any model is trained; Myth~\#1 rests on this fact.
Two surveys cover the field \citep{mielke2021between, cognetta2026tokenizationsurvey}.

Tokenization research asks how these choices affect a wide range of model properties.
A tokenizer that cuts text into fewer, longer tokens lowers the compute that a model spends per byte \citep{liu2025superbpe, pagnoni2024blt}.
The same content needs up to fifteen times more tokens in some languages than in others \citep{petrov2023unfairness}.
Speakers of those languages therefore pay more for commercial models and get poorer results \citep{ahia2023allcost}.
A monolingual tokenizer in place of a multilingual one raises a model's downstream scores on almost every task and language \citep{rust2021good}.
The tokenizer also decides what a model can do with the characters of its text.
How digits are split changes arithmetic accuracy \citep{singh2024tokenizationcounts}.
Models know the spelling of their tokens but fail at many manipulations of characters \citep{kaushal2022tokens, edman2024cute}, and byte-level models are more robust to noise \citep{xue2022byt5}.
In training, tokens that are almost absent from the data stay under-trained \citep{land2024magikarp}, and \citet{tao2024scalingvocab} predict that the vocabulary size that suits a model grows with its parameter count.
At inference, sampling the next byte from a tokenized model is biased unless its token probabilities are converted exactly to byte probabilities \citep{phan2025exact}.
A trained model can also be moved to a new tokenizer \citep{minixhofer2022wechsel, minixhofer2024zeroshot}.
Theory has followed these questions.
Greedy BPE approximates the best sequence of merges for compression: \citet{zouhar2023formal} tie its approximation factor to the unknown optimum, and \citet{kozma2026theoretical} bound the factor for every input.
Finding the tokenization that compresses a corpus most is NP-hard \citep{whittington2025tokenisation, lim2025partition}.
\citet{gastaldi2025foundations} give conditions under which a tokenizer keeps statistical estimators consistent.
On some Markov sources, transformers trained on characters fall back to a unigram model, while with a suitable tokenizer even a unigram model over tokens comes close to the optimal cross-entropy \citep{rajaraman2024analysis}.
In that setting the tokenizer, not the model, carries the structure of the source; Myth~\#1 measures the same kind of head start for a model that has learned nothing.
Most of these questions are answered by comparing models trained with different tokenizers, and such a comparison needs a metric that does not itself favor one tokenizer.
\Cref{app:related_comparing} reviews the work on such comparisons, after a short history of the tokenizers they compare.

\subsection{How Tokenizers Evolved}

\citet{sennrich2016neural} adapted byte pair encoding (BPE), a compression algorithm that repeatedly replaces the most frequent pair of adjacent symbols with a new symbol \citep{gage1994new}, to make translation open-vocabulary.
\citet{kudo2018subword} proposed the Unigram language model, which prunes a large vocabulary by likelihood; the unigram tokenizers of \Cref{app:tokeval_grid,app:decision_format} are of this kind.
Byte-level BPE starts from the 256 byte values, so every string has an encoding \citep{radford2019language, wang2020bytelevel}; the tokenizers of \Cref{fig:headline} are of this kind.
Most current open models use byte-level BPE, or a SentencePiece tokenizer that falls back to bytes \citep{kudo2018sentencepiece}, and their vocabularies have grown: GPT-2 had about 50K tokens \citep{radford2019language}, Llama~3 has 128K \citep{grattafiori2024llama3}, and Gemma~3 has 262K \citep{gemmateam2025gemma3}.
Recent tokenizers merge across pre-token boundaries into multi-word tokens \citep{liu2025superbpe, schmidt2025boundless}; \Cref{fig:superbpe} measures SuperBPE for untrained models.
A second line of work removes the fixed tokenizer, with models that read characters or bytes directly \citep{xue2022byt5} or pool bytes into segments \citep{yu2023megabyte, slagle2024spacebyte, pagnoni2024blt, hwang2026dynamic}.
These models share no vocabulary with subword models, so several of these papers compare the two in bits per byte (\Cref{sec:tokenizer_agnostic}).

\subsection{Comparing Models Across Tokenizers}
\label{app:related_comparing}

\paragraph{Units of comparison.}
A comparison across tokenizers has always needed a unit that belongs to the text.
\citet{shannon1951prediction} measured the entropy of English in bits per letter, and \citet{schuster2012japanese} noted that word perplexities cannot be compared across word inventories of different sizes.
Character-level models were scored in bits per character on enwik8 and text8, and several enwik8 results reported in bits per character are in bits per byte \citep{alrfou2019character}.
Models with different segmentations were placed on one scale by a conversion of units.
\citet{graves2013generating} converted bits per character to word perplexity through the average word length, and others converted bits per byte to word perplexity \citep{alrfou2019character, choe2019bridging}.
\citet{mielke2019comparing} stated the principle: perplexities per token cannot be compared across segmentations, but any metric whose denominator does not depend on the segmentation can.
This principle is property (P2) of \Cref{sec:impossibility}, and \Cref{thm:trilemma} shows that a metric with (P1) and (P2) gives up (P3).
Across languages the count of characters also depends on the writing system, so \citet{cotterell2018languages} divided by the characters of an aligned English translation.
GPT-2 divided the log-probability of each benchmark by that benchmark's own unit and reported bits per byte on enwik8 \citep{radford2019language}.
The Pile made bits per byte its preferred metric \citep{gao2020pile}, and Gopher used it to compare with models that have other tokenizers \citep{rae2021scaling}.
Byte-level models use it against subword baselines \citep{slagle2024spacebyte, pagnoni2024blt, hwang2026dynamic, lee2026efficiency}, and \Cref{sec:tokenizer_agnostic} quotes several of their reasons.
Two papers qualified this practice.
Paloma reports bits per byte but states that a normalization by bytes only partly corrects for different vocabularies, and that a fixed vocabulary is better \citep{magnusson2023paloma}.
\citet{deletang2024language} expected the two effects of a lossless tokenizer, fewer symbols and a larger alphabet, to compensate, and found instead that the compression rate of their transformers changed with the vocabulary.
Myth~\#1 gives a counterexample to automatic compensation: on English web text with BPE, the two effects do not cancel for a model that has learned nothing.

A separate line of work makes the probability of a string exact.
A string has many tokenizations, and its probability is the sum over all of them.
The canonical tokenization usually carries almost all of this sum, but not always on text outside the training domain or with long, complex words \citep{cao2021marginal, chirkova2023marginalize}.
Exact methods convert token probabilities into probabilities of words, characters or bytes \citep{pimentel2024word, vieira2024characters, phan2025exact, hayase2026sampling}, and a token model can be restricted to canonical tokenizations \citep{vieira2025canonical}.
\Cref{app:trilemma} defines the predictions of a pair through its canonical tokenization, the convention that these methods examine.

\paragraph{Tokenizer metrics and task scores.}
A statistic of a tokenizer's output that tracks task scores would let one choose a tokenizer without training a model on it.
Whether compression is such a statistic is disputed.
\citet{galle2019bpe} found in translation that, at a fixed vocabulary size, the compressor that needs fewer tokens gives the higher BLEU.
Lower fertility went with higher task scores across models with monolingual and multilingual tokenizers \citep{rust2021good}.
\citet{goldman2024unpacking} found that compression correlates with downstream performance when the tokenizer's training data, and with it the vocabulary size, is varied.
They read compression as a 0-gram language model that gives every token the same probability, which is the uniform model of \Cref{sec:tokenizer_agnostic}.
Their correlation is with the token count; the code length of that model is the token count times $\log V$, and $V$ varied across their tokenizers.
\citet{schmidt2024morethancompression} found no such relation across fifty-four runs of eighteen tokenizer variants; their token count correlates weakly with accuracy, with the sign opposite to the claim.
\citet{ali2024tokenizerchoice} report that fertility and parity ``are not always predictive of model downstream performance.''
These studies fix different things and score different outcomes, so their results do not conflict.
R\'enyi efficiency, another intrinsic metric, tracked translation quality in \citet{zouhar2023noiseless}, and \citet{cognetta2024counterexamples} built tokenizers that raise it while translation quality falls.
In recent studies, the tokenizer matters little for English tasks but consistently for multilingual ones \citep{lotz2025beyond}.
The sign of the correlation between token count and task score also changes with the kind of task \citep{wegmann2025variation}.

\paragraph{Controlled sweeps of tokenizers.}
Several studies train one recipe with several tokenizers and report that $\bpb$ separates the tokenizers more consistently than task scores do, and that the two rankings sometimes disagree.
At fixed model size and training compute, \citet{liu2025superbpe} report that ``the ranking of models according to BPB and downstream task performance are not consistent.''
\citet{land2026mingram} finds over twenty seeds that unigram-family tokenizers ``consistently beat BPE in bits-per-byte,'' while the task scores were ``too noisy at this scale to separate most methods.''
\citet{tempus2026convextok} find that their tokenizer ``consistently improves [...] the bits-per-byte (BpB) achieved by language models; it also improves downstream task performance, but less consistently.''
\citet{yavuz2026objective} find that bottom-up tokenizers ``consistently achieve lower bits-per-byte in most settings,'' while grammatical acceptability shows ``no consistent relationship between design choice and performance.''
\citet{meister2026tokeval} finds that information-theoretic tokenizer metrics correlate with $\bpb$, while metrics of structure, such as the handling of digits and line breaks, are the ones that correlate with task accuracy.
From byte-level and token-level models of about a billion parameters trained on up to a trillion bytes, \citet{marathe2026tokenceiling} conclude that ``BPB (Bits-Per-Byte) does not allow fair comparison across models with different training objectives (Distillation and Cross-Entropy) and tokenization schemes.''
They suggest that the gap may come from how much compute each model spends per unit of data at a fixed budget.
Of the five tokenizer sweeps, only \citet{liu2025superbpe} suggest a mechanism: the tokens a task scores tend to be among the hardest to predict.
\citet{marathe2026tokenceiling} also note that $\bpb$ reads only the probability of the correct target, while accuracy reads its rank.
\Cref{eq:answer-factorization} gives a general mechanism and makes this observation exact.
$\bpb$ charges for the probability outside the offered options and for exact rendering, and a change of tokenizer moves those two terms whether or not it moves the decision.
\Cref{sec:downstream} measures the split directly, on released models of \citet{altintas2026toksuite} and \citet{meister2026tokeval}.

\paragraph{The code length of the correct answer as a proxy for accuracy.}
Studies of scaling and of data selection use the code length of the correct answer as a smooth stand-in for accuracy.
Accuracy is discontinuous, and abilities that appear suddenly under accuracy change smoothly under continuous metrics \citep{schaeffer2023mirage}.
With one recipe and one tokenizer, models with the same pretraining loss reach the same task scores \citep{du2024loss}.
The average downstream error also falls predictably with the validation loss \citep{gadre2025overtraining}, although a reanalysis finds predictable scaling for fewer than half of the tasks \citep{lourie2025unreliable}.
Accuracy also depends on how the probability of each incorrect option changes \citep{schaeffer2024elusive}.
Current practice predicts the code length of the correct answer from training compute, or from model size and data, and maps it to accuracy with a sigmoid \citep{grattafiori2024llama3, bhagia2025ladders}.
\citet{bhagia2025ladders} divide this code length by bytes to reduce the effect of the tokenizer.
Small experiments that choose pretraining data rank the candidates by the probability or the bits per byte of the correct answer, because these are less noisy than accuracy \citep{magnusson2025datadecide, heineman2025signal}.
These studies mostly hold the tokenizer fixed.
Others compare across tokenizers.
Bits per character tracks benchmark scores across released models with different tokenizers \citep{huang2024compression}.
\citet{thrush2025perplexity} select pretraining data by the correlation between bits per byte and benchmark scores across ninety released models.
These correlations pool models that differ in size and training data as well as in the tokenizer; \Cref{sec:downstream} holds the layers and the training documents fixed and varies the tokenizer, and there the correlation is weak.
Loss-to-loss scaling laws change little with the tokenizer \citep{mayilvahanan2025line}.
\citet{dauncey2026tokenization} compare tokenization methods by the bits per byte of the correct continuation, because accuracy is at chance at their scale.
Their comparison ranks the methods by the code length of the correct continuations; it does not rank them by accuracy, and \Cref{fig:mcqa-continuation} shows that the two rankings can come apart.
A separate line of work studies the scored string.
When the answer text is scored, other strings with the same meaning compete for its probability \citep{holtzman2021surface}.
When the option letter is scored, the model must bind each letter to its option \citep{robinson2023leveraging}.
The two formats can give very different accuracies for the same model \citep{biderman2024lessons}, and the ability to answer with a letter appears abruptly during pretraining \citep{gu2025olmes, wiegreffe2025answer, bhagia2025ladders}.
\Cref{eq:answer-factorization} separates the two.

\paragraph{Reference models and vocabulary size.}
A second remedy divides out a reference model.
\citet{roh2020unigramnorm} divide perplexity by the perplexity of a unigram model.
Their proof splits one word type into two, and their experiments remove rare words from a word-level vocabulary.
These arguments cover changes to a word-level vocabulary; neither changes a subword segmentation.
\citet{tao2024scalingvocab} made this ratio the loss of their scaling laws for the vocabulary size, which \Cref{sec:unigram_normalized_loss} examines.
The vocabulary size itself has been tuned since subwords were introduced.
In translation, the best number of BPE merges depends on the architecture, and a poor choice alone can cost several BLEU points \citep{ding2019call}.
BLEU peaks at a vocabulary of 8K types or fewer on all but the largest data set of \citet{gowda2020finding}.
VOLT chooses the vocabulary by trading the entropy per character of a unigram token model against the vocabulary size \citep{xu2021volt}.
For large language models, \citet{tao2024scalingvocab} fit compute-optimal vocabulary sizes and predict that larger models need larger vocabularies.
Controlled sweeps report higher task scores with larger vocabularies \citep{takase2025large} and lower loss with larger input vocabularies \citep{huang2025overtokenized}, and the gain of a larger vocabulary is concentrated on the most frequent words \citep{chung2025exploiting}.
For code, the pre-tokenizer matters more than the vocabulary size \citep{dagan2024getting}.
\citet{limisiewicz2026computeoptimal} replace the vocabulary size with a compression rate and find that the loss-optimal rate lies near four bytes per token and falls slowly with compute.
To compare losses across vocabularies, three of these studies use a unit meant to remove the token: bits per byte \citep{limisiewicz2026computeoptimal}, a loss per word \citep{chung2025exploiting}, and the unigram-normalized loss \citep{tao2024scalingvocab}.
A loss per byte or per word includes the head start of Myth~\#1, which itself changes with the vocabulary.
The unigram-normalized loss is still a loss per token (\Cref{eq:unigram-normalized-loss-bpb}), and \Cref{sec:unigram_normalized_loss} shows that it ranks the same runs differently from bits per character.

\paragraph{Released sets of models trained with different tokenizers.}
At least four such sets exist.
The release that accompanies \citet{meister2026tokeval} holds 231 models over 98 tokenizers, with per-model $\bpb$ and task scores; \Cref{sec:downstream} uses the 102 runs that share the same layers and the same training documents (\Cref{app:tokeval_grid}).
\citet{altintas2026toksuite} release twenty-two models: one model of about a billion parameters for each of fourteen off-the-shelf tokenizers, all with the same number of training tokens, and eight more that vary the seed, the model size, or the amount of training text.
\citet{schmidt2024morethancompression} release the models of their study and ten larger ones, with per-model accuracy and no $\bpb$.
\citet{liu2025superbpe} release 8B models trained with BPE and with SuperBPE at the same training compute.

\section{Setup of the Vocabulary Sweeps}
\label{app:sweep-setup}

\paragraph{BPE on English.}
\Cref{fig:headline,fig:superbpe} use FineWeb-Edu-dedup \citep{penedo2024finewebdatasetsdecantingweb}, the \texttt{fineweb-edu-dedup} configuration of \texttt{HuggingFaceTB/smollm-corpus} at revision \texttt{3ba9d60}.
We split by parquet shard: the tokenizer trains on the leading shards and the held-out split, which every measurement reads, comes from the last eight, so no document is in both splits.
The training split holds $10.0$ GB and the held-out split $1.04$ GB.
Staging uses no random seed: it reads the shards in order and keeps every non-empty document until the split reaches its target size.
Each document ends with one newline, so the byte count $B$ is the file size.
We apply no Unicode normalization; the tokenizer reads raw UTF-8 bytes.
The tokenizer is byte-level BPE \citep{gage1994new, sennrich2016neural} with the GPT-2 pre-tokenization regex \citep{radford2019language} and no special tokens, trained with the Rust trainer of \texttt{gigatoken}.\footnote{\url{https://github.com/marcelroed/gigatoken}}
Its vocabulary is the $256$ byte values plus the merges, so $m$ merges give $V = 256 + m$.
We train one merge list to $V = 2^{20}$, and each smaller vocabulary is a prefix of it, so the pre-tokenizer and the training text are the same at every size.
BPE applies merges in rank order, so the number of times merge $j$ fires does not depend on any later merge.
One counting pass over the held-out pre-tokens therefore gives the exact token count $N_\tau$ at every integer vocabulary size from $256$ to $2^{20}$.
Every point of the BPE curves is observed, not interpolated or fitted, and each minimum is the smallest observed value.
Three checks confirm the counts: at $V = 256$ the count equals the byte count, at $2^{20}$ it equals a direct encode, and direct encodes with prefix tokenizers match it at several sizes in between.

\paragraph{Other corpora.}
The other corpora go through the same pipeline, each with its own merge list trained on its own training split.
Code is \texttt{codeparrot/github-code-clean} at revision \texttt{c48d40f}, restricted to twelve programming languages, with the last eight shards held out.
French, Hindi, Russian and Chinese are the \texttt{fra\_Latn}, \texttt{hin\_Deva}, \texttt{rus\_Cyrl} and \texttt{cmn\_Hani} partitions of FineWeb-2 at revision \texttt{af9c133}, with the last two shards of each partition held out.
Each training split holds $10.0$ GB.
Each held-out split holds $1.04$ GB, except French, whose last two shards hold only $0.223$ GB.
\Cref{tab:corpus-minima} lists the minimum of each sweep.

\paragraph{SuperBPE.}
\Cref{fig:superbpe} uses SuperBPE \citep{liu2025superbpe} with transition point $t = 2^{18}$.
Its first $t - 256$ merges are the English BPE merges.
Its later merges are learned on the same training split with no pre-tokenization, so a token can span whitespace and document boundaries.
We learn them in one greedy pass to $V = 2^{26}$, so each smaller SuperBPE vocabulary is a prefix.
Our implementation follows the released code of \citet{liu2025superbpe} at commit \texttt{bbd0976}, with three changes: our first stage uses the GPT-2 regex instead of theirs, our second stage drops pre-tokenization entirely where theirs keeps a reduced regex, and our first-stage merges never apply across a first-stage pre-token boundary.
We count SuperBPE tokens by encoding the held-out split directly on a grid of $67$ sizes from $256$ to $2^{26}$, a quarter octave apart above $2^{10}$, and every encoding above $t$ decodes back to the exact text.

\begin{table}[t]
\centering
\caption{\textbf{Myth \#1: Bits-Per-Byte Is Agnostic to the Tokenizer and/or Its Vocabulary Size.}
The minimum of $\bpb$ for untrained uniform models in each vocabulary sweep, on each corpus's held-out split. Each minimum is the smallest observed value: at every integer vocabulary size for BPE, and on the grid of $67$ sizes for SuperBPE. Chinese reaches its smallest value near the end of its sweep, where $\bpb$ is flat to six decimal places, so its $V$ is a lower bound and we count no minimum.
Where $\bpb$ is lowest for untrained models depends on the corpus and the tokenization algorithm.}
\label{tab:corpus-minima}
\begin{tabular}{lrrrr}
\toprule
 & & \multicolumn{2}{c}{At the minimum} & At the largest $V$ \\
\cmidrule(lr){3-4} \cmidrule(lr){5-5}
Corpus and tokenizer & Largest $V$ & $V$ & $\bpb$ & $\bpb$ \\
\midrule
English, BPE & $2^{20}$ & $24{,}114$ & $3.2386$ & $3.9320$ \\
Code, BPE & $2^{20}$ & $8{,}271$ & $4.2467$ & $5.3694$ \\
French, BPE & $2^{20}$ & $26{,}977$ & $3.3195$ & $3.9718$ \\
Hindi, BPE & $2^{18}$ & $504$ & $2.5790$ & $4.5190$ \\
Russian, BPE & $2^{20}$ & $117{,}360$ & $1.8260$ & $1.9632$ \\
Chinese, BPE & $2^{20}$ & $\geq 1{,}048{,}558$ & $3.2068$ & $3.2068$ \\
\midrule
English, SuperBPE ($t = 2^{18}$) & $2^{26}$ & $33{,}554{,}432$ & $2.0379$ & $2.0410$ \\
\bottomrule
\end{tabular}
\end{table}

\section{Why the Code Length Has a Minimum, and Where It Lies}
\label{app:downstream_derivation}

\paragraph{Under the uniform model, both metrics count tokens.}
Let $T_{i,c}$ be the number of tokens that option $c$ adds to the prompt of item $i$, and $\ell_{i,c}$ its length in characters.
Every token costs $\log V_\tau$ (\Cref{eq:nll-uniform}), so option $c$ costs $T_{i,c} \log V_\tau$, and
\begin{equation}\label{eq:mcqa-argmin}
\texttt{acc}:\;\; \arg\min_c T_{i,c},
\qquad
\texttt{acc\_norm}:\;\; \arg\min_c T_{i,c} / \ell_{i,c}.
\end{equation}
$\log V_\tau$ multiplies every option equally and cancels: vocabulary size reaches accuracy only through which option has the fewest tokens.
With the correct options as the corpus, \Cref{eq:bpb-uniform} gives
\begin{equation}\label{eq:bench-bpb}
\bpb(\tau, p^{\mathrm{unif}}) \;=\; \frac{N_\tau}{B} \, \log V_\tau ,
\end{equation}
and here $\log V_\tau$ does not cancel.
Tokens per byte fall as the vocabulary grows while each token gets more expensive.
These opposing trends permit a minimum; they do not force one.
On Chinese web text none appears up to $2^{20}$ (\Cref{tab:corpus-minima}).

\paragraph{Where the minimum lies.}
For two tokenizers with $V_{\tau_1} < V_{\tau_2}$, \Cref{eq:bench-bpb} gives
\begin{equation}\label{eq:bench-discrete}
\bpb(\tau_2, p^{\mathrm{unif}}) < \bpb(\tau_1, p^{\mathrm{unif}})
\quad\Longleftrightarrow\quad
\frac{N_{\tau_2}}{N_{\tau_1}} < \frac{\log V_{\tau_1}}{\log V_{\tau_2}}.
\end{equation}
$\bpb$ falls exactly when the larger vocabulary cuts the token count by a larger factor than it raises the cost of each token.
A label stops shortening once it is a single token.
MMLU's four labels become single tokens at $V_\tau = 327$, $336$, $380$ and $413$; at $413$, tokens per byte reach exactly $\tfrac12$ and $\bpb$ reaches its minimum, $4.34$.
Above that, $\bpb$ is exactly $\tfrac12 \log_2 V_\tau$.
Across the five letter benchmarks the minima lie between $361$ and $484$.
On MMLU, ARC-Easy and HellaSwag, answer text keeps shortening, so its minimum lies four to seven octaves higher.
The matched pairs show this directly: from labels to answer text, the minimum moves from $413$ to $38{,}968$ on MMLU, to $23{,}170$ on ARC-Easy, and to $11{,}585$ on HellaSwag.
What sets the location is how fast merges shorten the strings, not their length: GPQA diamond's answers are as long as MMLU's ($43.8$ against $44.3$ bytes), but they shorten slowly, and their minimum moves only from $361$ to $435$.

\section{Formal Statement of the Tokenizer Trilemma}
\label{app:trilemma}

\paragraph{Setting.}
Fix the finite set $\mathcal{X}$ of all texts of $n$ bytes.
Write $\tau(x) = (t_1, \dots, t_{N_\tau(x)})$ for the canonical tokenization of $x$, the tokens that the encoder of $\tau$ outputs.
A pair $(\tau, p)$ gives the canonical path $\tau(x)$ the mass
\begin{equation}\label{eq:forecast}
    q_{\tau,p}(x) \;=\; \prod_{k=1}^{N_\tau(x)} p\big(t_k \mid t_{<k}\big),
\end{equation}
and $\bpb$ as computed in practice scores this mass: $\bpb(\tau, p) = \frac{1}{B} \sum_{d} -\log q_{\tau,p}(x^{(d)})$.
The masses sum over $\mathcal{X}$ to some $m_{\tau,p} \le 1$, because $p$ can also spend probability on token sequences that are not canonical tokenizations of texts in $\mathcal{X}$.
So $q_{\tau,p}$ is a distribution only when $m_{\tau,p} = 1$; the normalized distribution of the pair is $Q_{\tau,p} = q_{\tau,p} / m_{\tau,p}$.
A metric assigns a loss $S(\tau, p; x)$ to a pair on a text, lower being better.
Formally, the three properties are as follows:
\begin{itemize}[leftmargin=2.6em, itemsep=1pt, topsep=2pt]
    \item[\textbf{(P1)}] \emph{Only the truth scores best} (strict propriety; \citealp{gneiting2007proper}). For every distribution $P$ on $\mathcal{X}$ and every $\tau$: $\E_P\, S(\tau, p; X) < \E_P\, S(\tau, p'; X)$ whenever $q_{\tau,p} = P \neq q_{\tau,p'}$.
    \item[\textbf{(P2)}] \emph{Same predictions, same score} (representation invariance). If $q_{\tau,p} = q_{\tau',p'}$, then $S(\tau, p; x) = S(\tau', p'; x)$ for every $x$.
    \item[\textbf{(P3)}] \emph{Untrained models tie} (null neutrality). Each $\tau$ has a designated reference model $p^0_\tau$, and $\E_P\, S(\tau, p^0_\tau; X) = \E_P\, S(\tau', p^0_{\tau'}; X)$ for every $P$ and every $\tau, \tau'$.
\end{itemize}
(P1) and (P2) are stated for the masses that $\bpb$ scores, and the per-text $\bpb$, $-\log q_{\tau,p}(x) / n$, has both.
The data distribution $P$ is normalized, and so are the masses of the references and of the copy in the proof below; for these models, $q_{\tau,p} = Q_{\tau,p}$.
The theorem does not by itself describe every empirical quantity called $\bpb$, such as one over texts of different lengths, or one that sums the probability of a text over all of its tokenizations.
(P3) asks for a tie under every $P$ because neutrality should not depend on which corpus is evaluated; the proof uses one $P$ only.

\paragraph{Corpus ratios and per-text averages.}
(P1) classifies a score by its value under $P$.
For a per-text loss, this value is $\E_P\, S$.
For a corpus ratio of total log loss to total length, it is the large-corpus limit $\E_P[-\log q_{\tau,p}(X)] / \E_P[\ell_\tau(X)]$, where $\ell_\tau(x)$ is the length of $x$ in bytes or in tokens of $\tau$.
With the tokenizer fixed, this denominator does not depend on $p$, so the ratio keeps (P1), and the proof below holds with this limit in place of $\E_P\, S$.
An average of per-text losses, each divided by its own length, is minimized by the distribution proportional to $P(x) / \ell_\tau(x)$, so it breaks (P1) unless every text has the same length.
On $\mathcal{X}$ the two forms coincide for bytes, but not for tokens: $N_\tau(x)$ varies among texts of $n$ bytes.
The unigram-normalized loss (\Cref{eq:unigram-normalized-loss}) is a corpus ratio over tokens.
With the tokenizer fixed, its unigram term and its denominator do not depend on $p$, so it keeps (P1).
Each tokenizer's unigram model scores zero, so it keeps (P3) with that model as the reference.
Pairs with the same predictions under different tokenizers get different unigram terms and denominators, so it gives up (P2).

\paragraph{Representability.}
Under a lossless tokenizer, every distribution $Q$ on $\mathcal{X}$ is the mass of some model: let $p(t_k \mid t_{<k})$ be the $Q$-mass of the texts whose canonical tokenization begins with $t_{\le k}$, divided by that of the texts whose canonical tokenization begins with $t_{<k}$.
No canonical tokenization of an $n$-byte text is a proper prefix of another, because a proper prefix spells fewer bytes, so \Cref{eq:forecast} telescopes to $q_{\tau,p}(x) = Q(x)$.
This construction uses arbitrary conditional distributions, so the theorem is an impossibility result for scores over the unrestricted class of autoregressive models.
A fixed finite architecture need not represent every distribution; for such a class, the argument needs the class to contain the references and the copy.

\paragraph{The uniform reference.}
The uniform model's mass $V_\tau^{-N_\tau(x)}$ sums to some $m_\tau \le 1$ over $\mathcal{X}$, because the model also spends probability on token sequences that are not canonical tokenizations of texts in $\mathcal{X}$.
We therefore take as reference the uniform model conditioned on emitting a canonical tokenization of a text in $\mathcal{X}$, whose mass is the normalized distribution $Q^0_\tau(x) = V_\tau^{-N_\tau(x)} / m_\tau$.
Conditioning adds the same constant, $\log m_\tau$, to the code length of every text, so it cannot remove a head start that varies from text to text.
The constant differs between tokenizers, so this reference is not identical to the raw uniform model of \Cref{fig:headline}, and the two can rank tokenizers differently.
Two uniform references coincide only if $N_\tau(x) \log V_\tau$ differs between the tokenizers by one constant across $\mathcal{X}$; BPE and the byte tokenizer fail this as soon as two $n$-byte texts receive different numbers of BPE tokens.

\begin{theorem}[\Cref{thm:trilemma}, formal]
Let $\tau_1$ and $\tau_2$ be lossless tokenizers on $\mathcal{X}$ whose references predict distinct distributions, $Q^0_{\tau_1} \neq Q^0_{\tau_2}$.
No metric on pairs built from $\tau_1$ and $\tau_2$ has (P1), (P2), and (P3).
\end{theorem}

\begin{proof}
Let $P = Q^0_{\tau_1}$, and let $p'$ be a model with $q_{\tau_1,p'} = Q^0_{\tau_2}$, which exists by representability.
Then
\begin{equation*}
    \E_P\, S(\tau_1, p^0_{\tau_1}; X)
    \;\overset{\text{(P1)}}{<}\;
    \E_P\, S(\tau_1, p'; X)
    \;\overset{\text{(P2)}}{=}\;
    \E_P\, S(\tau_2, p^0_{\tau_2}; X),
\end{equation*}
so the two references do not tie under $P$, which contradicts (P3).
\end{proof}

\section{Table~2 of \texorpdfstring{\citet{tao2024scalingvocab}}{Tao et al. (2024)} Does Not Support Their Law}
\label{app:tao_table2}

\begin{figure}[h]
\centering
\includegraphics[width=\textwidth]{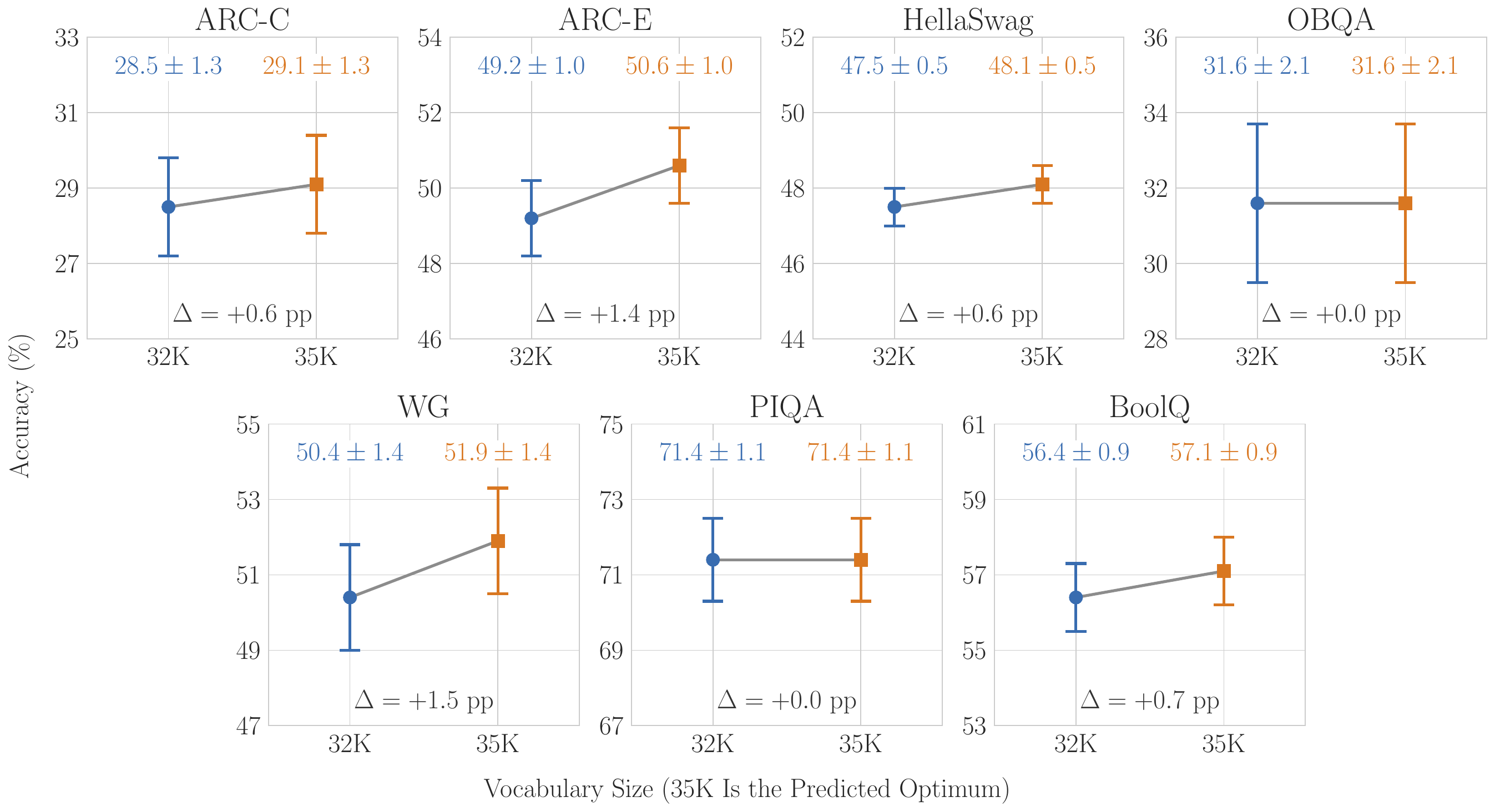}
\caption{\textbf{Table~2 of \citet{tao2024scalingvocab}, With Its Reported Error Bars.}
Each panel is one benchmark, and each point is one trained model.
The bars are the reported $\pm$ values, which are numerically consistent with the standard errors of the mean over test items.}
\label{fig:tao-table2}
\end{figure}

\citet{tao2024scalingvocab} support their predicted vocabulary with a downstream comparison in their Table~2.
Two models with $2.87$B non-vocabulary parameters, trained with the same $1.2 \times 10^{21}$ FLOPs, differ in vocabulary: $32$K, and the predicted optimum, $35$K.
At this fixed budget, the vocabulary also changes how many tokens each model trains on and how many characters of text those tokens cover.
The $35$K model's reported average accuracy over seven benchmarks is $0.6$ points higher (\Cref{fig:tao-table2}).
This gain does not support the law that predicts it, for three independent reasons: two concern the statistics, and one concerns the released fitting code.

\textbf{The law says the two models are the same.}
Evaluated at this setting, the released loss model of their Approach~3 predicts values of $\mathcal{L}_u$ for the two vocabularies that differ by $4.5 \times 10^{-5}$ nats per token: the changes in its vocabulary term and its data term almost cancel.
For scale, $\mathcal{L}_u$ in their largest released training run scatters by about $0.009$ nats per token around its trend from one checkpoint to the next.
A change this small in the mean loss does not tightly bound the change in accuracy, because probability can move between test items with little change in the mean; items near the decision boundary are one way this can happen.
So the law does not rule out a gain of a point, but it gives no reason to expect one.

\textbf{The accuracy gain is uncertain.}
Each vocabulary is one trained model.
The $\pm$ values in their Table~2, which the caption calls standard deviations, are numerically consistent with standard errors of the mean over test items: they match the formula of Lighteval, the evaluation library the paper uses.
No per-task gain exceeds the standard error of an unpaired difference, $\sqrt{2}$ times the reported value.
An unpaired test is not the right one, because both models answer the same items.
The paper releases no per-item outputs and no repeated training runs, so the released summaries do not establish a reproducible accuracy advantage or permit the preferred paired analysis.

\textbf{The predicted vocabulary comes from a units error.}
The released fitting code converts characters to tokens by evaluating the tokens-per-character function at $V/1000$ instead of $V$.
We refit the law with the released procedure: L-BFGS-B minimizes a Huber loss from each of $486$ starting points, and the fit kept is the one with the lowest relative squared error among those whose two exponents lie between $0.1$ and $1$.
Refit with each run's actual token count, the same law describes the training losses about as well ($R^2 = 0.968$, against $0.972$), but its optimum at this budget moves to about $20$K, and it predicts that $35$K is slightly worse than $32$K.
Similar $R^2$ values do not locate this optimum: refits that hold it anywhere from $10$K to $64$K reach an $R^2$ within $0.001$ of the best.
It stays below $32$K, though: with the actual token counts, the released procedure selects an optimum below $32$K on two computing platforms and in each of $60$ refits that leave out one training run.
Each of $2{,}486$ starting points predicts that $35$K is worse than $32$K.
The code, both conversions, the objective, the starting points, the selected coefficients, and these checks are in \texttt{results/reanalysis/} in the code repository, which will be released.

\section{TokEval's Released Models: Task Score Against Code Length on the Benchmark's Own Text}
\label{app:tokeval_grid}

\paragraph{Models.}
\citet{meister2026tokeval} released 231 trained models with their scores on BLiMP, GSM8K, MBPP, and HumanEval, in the Hugging Face repository \texttt{cmeister/tokenizer-lm-ablations} (revision \texttt{da6f830b}); where TokEval's paper and this release differ, we follow the release.
We use the \texttt{std-1B} regime, in which every model has the same 24 layers of width $1{,}536$, a context of $2{,}048$ tokens, and a batch of $2^{20}$ tokens per step, and saw the same documents in the same order with the same optimizer settings.
The tokenizer changes four other quantities.
The vocabulary, 64K to 200K entries, sets the size of the embedding tables, so the parameter count is 0.98B to 1.60B.
The same documents take 9.0B to 10.5B tokens, so the number of training steps changes too.
With the tokens per byte, the bytes in one context and the compute spent on each byte change as well.
After excluding six runs whose names mark continued or token-matched training, 102 runs remain over 85 tokenizers.
We average a tokenizer's repeated runs.

\paragraph{Text.}
The released evaluation code scores the models with tasks of the lm-evaluation-harness \citep{eval-harness}, and we take every string from the same tasks.
GSM8K has 1,319 test problems, each after the eight fixed chain-of-thought examples of its task.
MBPP has 500 test problems, each after three fixed examples.
HumanEval has 164 problems and no examples.
BLiMP has 67,000 sentence pairs and no prompt.
For each item, the scored text is the problem followed by its reference answer: the reference solution for GSM8K, the reference code for MBPP, the canonical solution for HumanEval, and the grammatical sentence of the pair for BLiMP, which has no problem.
The few-shot examples condition the model and are not charged, because they are the same for every item.
We write each GSM8K reference solution in the format of the task's examples: the calculator annotations are removed, the steps are joined with one space, and the text ends with ``The answer is $n$.''
Each text follows the beginning-of-sequence token, as in training.
$\bpb$ is the code length of the scored text, summed over the items of a benchmark, divided by its bytes.

\paragraph{Checks.}
One of the 102 models cannot be scored, because its tokenizer needs a loader that is not public.
For each of the other 101 models we recompute BLiMP accuracy from our own log-likelihoods of the two sentences of each pair.
It agrees with the released accuracy within $0.0013$ for every model, so the models and tokenizers load as they did in the original evaluation.

\paragraph{Intervals.}
Every interval of \Cref{fig:tokeval-grid}, of its discussion in \Cref{sec:downstream}, and of this appendix is a $95\%$ percentile bootstrap interval over tokenizers.
The unit is the tokenizer: the runs of one tokenizer are averaged first, so each tokenizer is one point however often it was trained.
Each interval draws $10{,}000$ resamples of the tokenizers with replacement, each tokenizer keeping its code length and score together.
The Pearson and the Spearman correlation are computed on the same resamples, and the interval runs from the 2.5th to the 97.5th percentile.
The resampling treats the tokenizers as independent, although many share their training data or their algorithm.
It does not carry the uncertainty from the training runs or from the benchmark items: each tokenizer's code length and score enter as fixed values.
The band around each least-squares line is a percentile interval of the line refitted on $1{,}000$ resamples of the same points.

\begin{figure}[t]
\centering
\includegraphics[width=\textwidth]{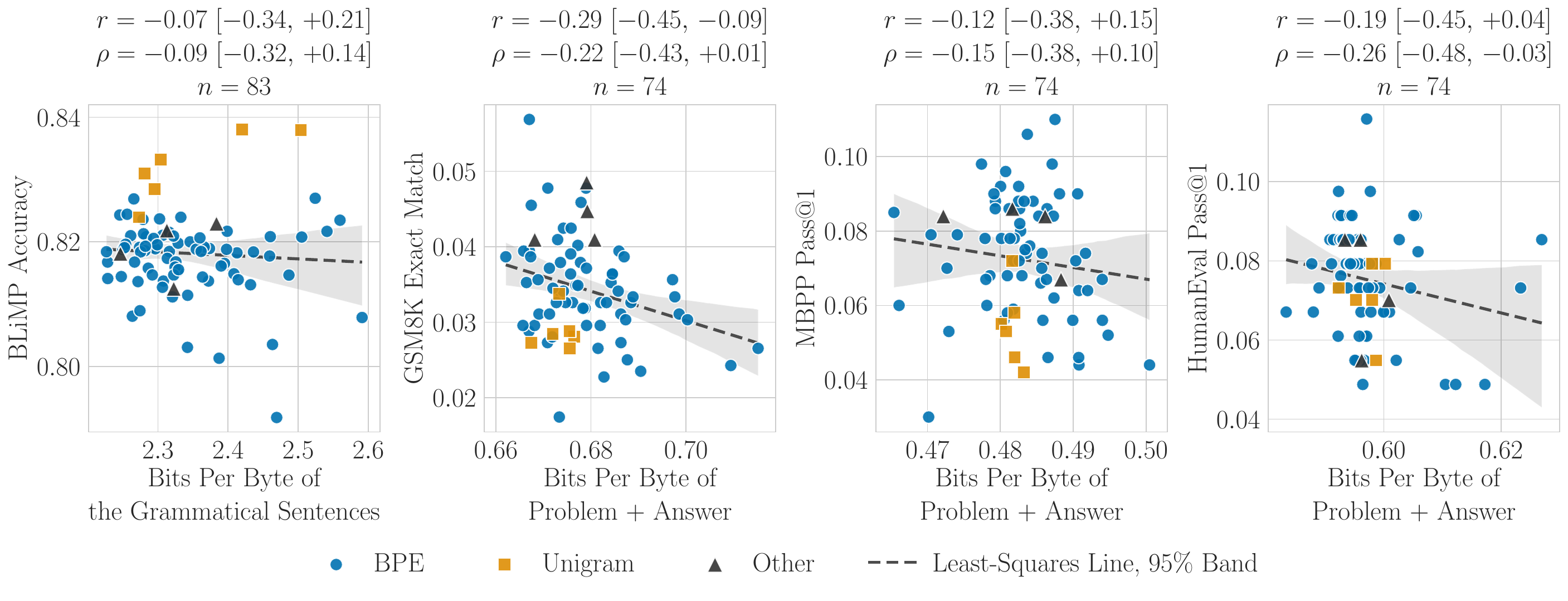}
\caption{\textbf{\Cref{fig:tokeval-grid} Without the Outlier Tokenizer.}
Axes, markers, and statistics as in \Cref{fig:tokeval-grid}, now over 83 scored tokenizers (74 for the three generated tasks), without \texttt{whitespace-balanced-unigram}.
Without the outlier, every correlation stays weak.}
\label{fig:tokeval-grid-without-outlier}
\end{figure}

\paragraph{One outlier tokenizer.}
\Cref{fig:tokeval-grid} and its discussion in \Cref{sec:downstream} include every scored tokenizer; the red cross is \texttt{whitespace-balanced-unigram}, an outlier.
It is a unigram tokenizer whose pre-tokenizer drops one space at the start of a text.
Its held-out $\bpb$ in the released grid is the highest by a wide margin, $0.840$ against $0.749$ for the next tokenizer, and its scores are $0.030$ on GSM8K, $0.040$ on MBPP, and exactly zero on HumanEval.
\Cref{fig:tokeval-grid-without-outlier} repeats \Cref{fig:tokeval-grid} without it: the Pearson correlations move by at most $0.093$, and every correlation stays weak.
The noise ceilings, \Cref{tab:heldout-signal}, and the held-out comparisons below leave it out; \Cref{fig:decision-share} and \Cref{app:decision_format} keep it among their TokEval tokenizers.

\paragraph{Noise.}
The released scores are noisy.
From 17 configurations that \citet{meister2026tokeval} ran twice (15 for GSM8K, MBPP, and HumanEval), the standard deviation of a score across tokenizers is $2.4$ times its run-to-run standard deviation for BLiMP, $1.4$ times for GSM8K, $1.2$ times for MBPP, and $1.6$ times for HumanEval.
If the noise is independent and additive, a score that spreads $k$ times its run-to-run standard deviation can correlate with no quantity more strongly than $\sqrt{1 - k^{-2}}$.
With the run-to-run noise of the code length as well, measured on the same 17 configurations, the ceilings are $0.90$ for BLiMP, $0.66$ for GSM8K, $0.51$ for MBPP, and $0.77$ for HumanEval.
With the outlier tokenizer or without it, every Pearson correlation of \Cref{fig:tokeval-grid} is less than half its ceiling, so noise weakens these correlations but does not explain why they are weak.

\begin{table}[t]
\centering
\caption{\textbf{The Sign of a Correlation Between a Code Length and a Score Depends on the Text the Code Length Is Measured On.}
Spearman correlation across TokEval's tokenizers between a released $\bpb$ and a released score, with $95\%$ bootstrap intervals; $n = 84$ for BLiMP and $74$ for the other three.
A negative sign means that the shorter code goes with the higher score.
The held-out mixture is TokEval's own validation text; the code and multilingual texts are its released code and FLORES evaluations.}
\label{tab:heldout-signal}
\vspace{4pt}
\footnotesize
\begin{tabular}{@{}lcccc@{}}
\toprule
$\bpb$ measured on & BLiMP & GSM8K & MBPP & HumanEval \\
\midrule
Held-out mixture & \begin{tabular}[c]{@{}c@{}}$+0.11$\\$[-0.13, +0.34]$\end{tabular} & \begin{tabular}[c]{@{}c@{}}$-0.65$\\$[-0.78, -0.47]$\end{tabular} & \begin{tabular}[c]{@{}c@{}}$-0.16$\\$[-0.39, +0.09]$\end{tabular} & \begin{tabular}[c]{@{}c@{}}$-0.26$\\$[-0.45, -0.05]$\end{tabular} \\[2pt]
Code & \begin{tabular}[c]{@{}c@{}}$+0.17$\\$[-0.06, +0.38]$\end{tabular} & \begin{tabular}[c]{@{}c@{}}$-0.59$\\$[-0.74, -0.39]$\end{tabular} & \begin{tabular}[c]{@{}c@{}}$-0.18$\\$[-0.40, +0.06]$\end{tabular} & \begin{tabular}[c]{@{}c@{}}$-0.31$\\$[-0.48, -0.11]$\end{tabular} \\[2pt]
Multilingual & \begin{tabular}[c]{@{}c@{}}$+0.31$\\$[+0.10, +0.48]$\end{tabular} & \begin{tabular}[c]{@{}c@{}}$-0.14$\\$[-0.37, +0.09]$\end{tabular} & \begin{tabular}[c]{@{}c@{}}$+0.36$\\$[+0.15, +0.54]$\end{tabular} & \begin{tabular}[c]{@{}c@{}}$-0.04$\\$[-0.26, +0.18]$\end{tabular} \\
\bottomrule
\end{tabular}
\end{table}

\paragraph{Held-out text as an early signal.}
A reply to \Cref{fig:tokeval-grid}, made by \citet{limisiewicz2026bpbthread}, is that $\bpb$ belongs not on the benchmark's own text but on a held-out slice of the pretraining mixture, where it is ``a good early signal of LM performance that allows meaningful comparison across tokenizers,'' though ``not a perfect predictor.''
This reply compares a code length on one distribution with a score on another, and a correlation between the two, of either sign, is not evidence that the code length measures the score.
TokEval's held-out mixture is 37\% English web text, 33\% multilingual web text, 16\% mathematics, and 14\% code by bytes \citep[Table~7]{meister2026tokeval}; GSM8K is grade-school math word problems.
On these models the sign of such a correlation depends on which text is chosen (\Cref{tab:heldout-signal}).
$\bpb$ on the held-out mixture co-ranks with GSM8K, Spearman $-0.65$ $[-0.78, -0.47]$ ($n = 74$), while $\bpb$ on multilingual text co-ranks with BLiMP and MBPP in the wrong direction: the tokenizers with the longer multilingual code have the higher scores.
Where the differences are small, between tokenizers whose GSM8K scores differ by less than one run-to-run standard deviation, the lower held-out $\bpb$ goes with the higher score in 61\% of pairs, against 50\% for a coin flip.
The comparison of like with like, $\bpb$ on the GSM8K text against the GSM8K score, gives only $-0.22$ $[-0.43, +0.01]$ over the same 74 tokenizers.
Whatever the held-out $\bpb$ tracks on GSM8K, it is not how well the model codes GSM8K text.
The comparisons of this paper therefore measure the code length and the score on the same strings.

\section{The Format and Decision Terms on Trained Models}
\label{app:decision_format}

\paragraph{Models.}
TokSuite \citep{altintas2026toksuite} trained one network of about 1B non-embedding parameters with 14 tokenizers, from bytes to a 256K vocabulary, on 100B tokens each.
We use 16 of its released models: the 14 main models, one per tokenizer, and two more seeds of the Llama~3.2 model, which with its main model give three runs that differ in the model seed only.
TokEval (\Cref{app:tokeval_grid}) fixes the documents and their order; we use 33 of its \texttt{std-1B} runs over 24 tokenizers: the vocabulary sweep, the unigram tokenizers and the BPE tokenizers matched to them, the Llama~3 tokenizer, the tokenizers trained on English or code data, and the three at the extremes of held-out $\bpb$; 9 configurations have a second seed.
Each model is scored on four benchmarks, MMLU, ARC-Easy, HellaSwag, and GPQA, once on the option letter and once on the answer text, with the lm-evaluation-harness \citep{eval-harness}: one string per option, the beginning-of-sequence token in front as in training, and the model in float32.
An option $y$ after a prompt $x$ is scored as in TokSuite's own evaluation: the model reads the tokens of the whole string $xy$, and the option is charged for the tokens from position $|\mathrm{enc}(x)|$ on.
The harness's own rule feeds $\mathrm{enc}(x)$ followed by those tokens, a sequence that the tokenizer never produces when $\mathrm{enc}(x)$ is not a prefix of $\mathrm{enc}(xy)$.
This happens only for two TokSuite tokenizers, TokenMonster and BLOOM; for every other tokenizer the two rules give the same numbers.
Under either rule, the tokens charged to the option can then miss some of its bytes or include some of the prompt's.
\Cref{tab:decision-slopes} gives the slopes under both rules.

\paragraph{Terms.}
For an item with prompt $x$, scored strings $y_1, \dots, y_k$ and correct index $g$, let $p(y_c \mid x)$ be the harness's score of option $c$: the product of the model's probabilities of the tokens charged to it.
The format term is $-\log_2 \sum_c p(y_c \mid x)$ and the decision term is $-\log_2 \big[ p(y_g \mid x) / \sum_c p(y_c \mid x) \big]$.
Their sum is the code length of the correct string, checked per item.
Here $F$ is the event that the response is one of the $k$ scored strings, so a correct paraphrase lies outside $F$.
The rendering term of \Cref{eq:answer-factorization} is zero under this definition of $F$, because an option is one string.
The split is algebraic: it holds for the harness's scores, which are not exact probabilities of the scored strings, because the harness appends no end symbol and scores one tokenization of each string.
Per tokenizer and benchmark, each term is summed over the items and divided by the bytes of the correct strings; the runs of one tokenizer are averaged.
\Cref{fig:decision-share} gives the slope of the decision term $D$ on the $\bpb$ $B$ of the correct answers across tokenizers, $\operatorname{Cov}(D, B) / \operatorname{Var}(B)$.
The format term's slope is one minus it, because $B$ is the sum of the two terms.
A slope is a least-squares coefficient, not a fraction of the differences: it falls below zero when the decision term moves against $B$, and above one when the format term does.
The decision term's fraction of the $\bpb$ itself, which \Cref{sec:downstream} quotes for MMLU letters, is the mean decision term over the mean $\bpb$.

\begin{table}[t]
\caption{\textbf{The Decision Term's Slope in All Sixteen Conditions, Under Both Scoring Rules.}
Slope: $\operatorname{Cov}(D, B) / \operatorname{Var}(B)$ across tokenizers, with its $95\%$ bootstrap interval.
\Cref{fig:decision-share} draws all sixteen under TokSuite's rule; the harness's rule changes only TokSuite's numbers.}
\label{tab:decision-slopes}
\centering
\small
\begin{tabular}{@{}llr@{\hspace{1.2em}}r@{}}
\toprule
& & \multicolumn{1}{c}{TokSuite's Rule (Used)} & \multicolumn{1}{c}{Harness's Rule} \\
\cmidrule(lr){3-3} \cmidrule(l){4-4}
Benchmark & Scored & Slope $[95\%$ Interval$]$ & Slope $[95\%$ Interval$]$ \\
\midrule
\multicolumn{4}{@{}l}{\textit{TokSuite, 14 tokenizers}} \\
MMLU & Letter & $-0.01$ $[-0.11, +0.04]$ & $+0.04$ $[-0.10, +0.04]$ \\
MMLU & Text & $+0.33$ $[+0.25, +0.49]$ & $+0.19$ $[+0.15, +0.49]$ \\
ARC-Easy & Letter & $+0.04$ $[+0.01, +0.08]$ & $+0.04$ $[+0.02, +0.07]$ \\
ARC-Easy & Text & $+0.04$ $[-0.12, +0.17]$ & $+0.06$ $[-0.08, +0.17]$ \\
HellaSwag & Letter & $+0.18$ $[+0.15, +0.23]$ & $+0.08$ $[+0.05, +0.23]$ \\
HellaSwag & Text & $+0.29$ $[+0.18, +0.59]$ & $+0.26$ $[+0.17, +0.37]$ \\
GPQA & Letter & $-0.07$ $[-0.26, +0.03]$ & $-0.07$ $[-0.26, +0.03]$ \\
GPQA & Text & $+0.19$ $[+0.11, +0.25]$ & $+0.17$ $[+0.11, +0.21]$ \\
\midrule
\multicolumn{4}{@{}l}{\textit{TokEval, 24 tokenizers}} \\
MMLU & Letter & $+0.05$ $[-0.04, +0.08]$ & \multicolumn{1}{c}{Same} \\
MMLU & Text & $+0.08$ $[+0.01, +0.18]$ & \multicolumn{1}{c}{Same} \\
ARC-Easy & Letter & $+0.07$ $[+0.00, +0.11]$ & \multicolumn{1}{c}{Same} \\
ARC-Easy & Text & $+0.03$ $[-0.01, +0.07]$ & \multicolumn{1}{c}{Same} \\
HellaSwag & Letter & $+0.15$ $[+0.04, +0.27]$ & \multicolumn{1}{c}{Same} \\
HellaSwag & Text & $+0.46$ $[+0.38, +0.57]$ & \multicolumn{1}{c}{Same} \\
GPQA & Letter & $+0.03$ $[-0.02, +0.08]$ & \multicolumn{1}{c}{Same} \\
GPQA & Text & $+0.25$ $[+0.15, +0.39]$ & \multicolumn{1}{c}{Same} \\
\bottomrule
\end{tabular}

\end{table}

\begin{figure}[p]
\centering
\includegraphics[width=0.90\textwidth]{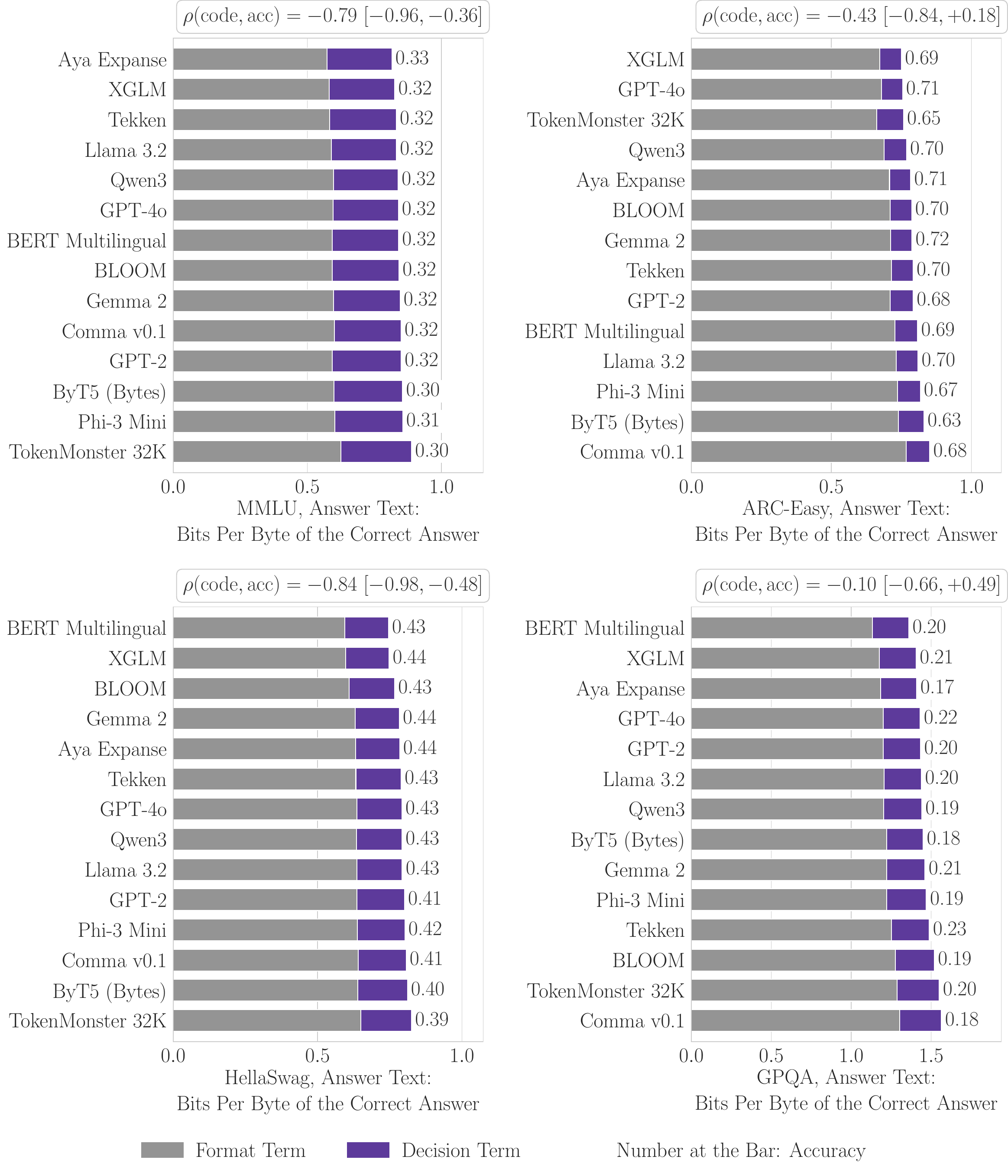}
\caption{\textbf{TokSuite: The Format and Decision Terms of Every Tokenizer.}
One bar per tokenizer on the answer text of each benchmark, sorted by the code length of the correct answer: the format term, then the decision term, with the accuracy at the end of the bar.
The box gives the Spearman correlation of the code length with accuracy across tokenizers and its $95\%$ bootstrap interval.}
\label{fig:terms-toksuite}
\end{figure}

\begin{figure}[p]
\centering
\includegraphics[width=0.90\textwidth]{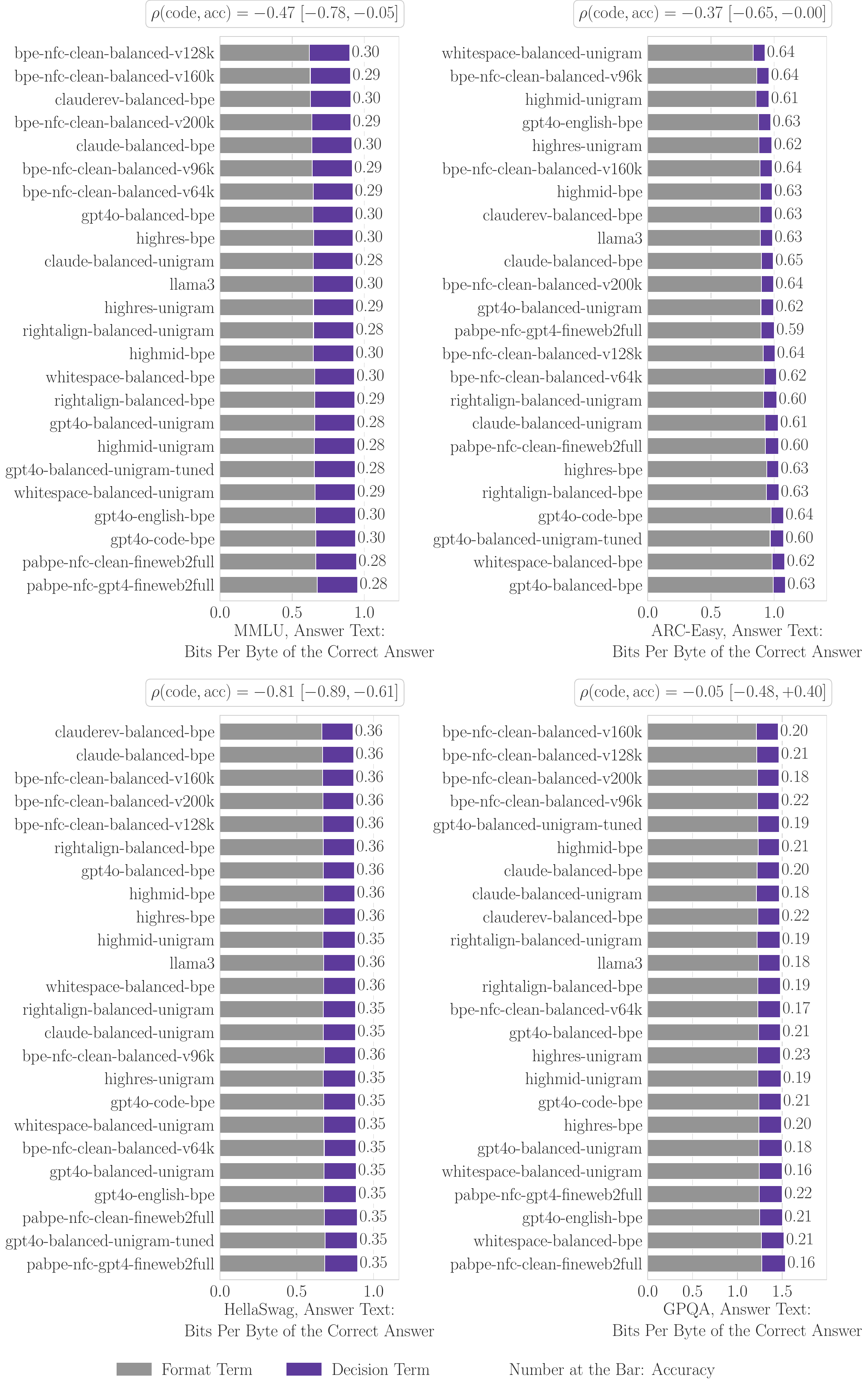}
\caption{\textbf{TokEval: The Format and Decision Terms of Every Tokenizer.}
As \Cref{fig:terms-toksuite}, for the 24 TokEval tokenizers, which saw the same documents in the same order; runs of one tokenizer are averaged.}
\label{fig:terms-tokeval}
\end{figure}

\paragraph{Results.}
\Cref{fig:terms-toksuite,fig:terms-tokeval} give the two terms and the accuracy of every tokenizer on the answer text of each benchmark.
In TokEval, which fixes the documents, accuracy co-ranks across the $24$ tokenizers with the decision term at $\rho = -0.94$ $[-0.98, -0.83]$ on HellaSwag and $-0.89$ $[-0.97, -0.70]$ on ARC-Easy, and with the code length of the correct answer at $-0.81$ $[-0.89, -0.61]$ and $-0.37$ $[-0.65, -0.00]$.
On MMLU, the two are alike: $-0.43$ $[-0.74, +0.01]$ for the decision term and $-0.47$ $[-0.78, -0.05]$ for the code length.

\paragraph{Why HellaSwag differs.}
Only on HellaSwag answer text does the interval of the decision term's slope reach above one half (\Cref{tab:decision-slopes}).
HellaSwag's correct endings were written by people, and its wrong endings are samples from a GPT language model, filtered so that a BERT model found them hard to tell from the real ones \citep{zellers2019hellaswag}.
The models are pretrained on mostly human-written text.
A pair that codes such text in fewer bits per byte codes the human-written correct ending in fewer bits but gains less on the machine-written wrong endings, so the decision term falls with the $\bpb$.
Most of the slope remains without the passage: scored after the beginning-of-sequence token alone, the decision term keeps $95\%$ $[65\%, 144\%]$ of its slope across TokEval's tokenizers.
Without the passage, the models pick the correct ending $31\%$ of the time, against $25\%$ by chance and $36\%$ with the passage.

\section{Randomly Initialized Models Are Approximately Uniform Under Modern Parameterizations}
\label{app:random-init}

The untrained models of this paper give every token of their vocabulary the same probability, $p^{\mathrm{unif}}(t) = 1/V_\tau$.
Under $\mu$P and its successors, the parameterizations used to transfer hyperparameters from small models to large ones, a randomly initialized model is close to $p^{\mathrm{unif}}$, and it comes closer as its width grows.

\paragraph{The logit scale sets the distance from uniform.}
At a fixed context, a model of width $w$ with vocabulary size $V = V_\tau$ has logits $z = \alpha W h$, where $h \in \mathbb{R}^w$ is the output of the final normalization layer, $W \in \mathbb{R}^{V \times w}$ is the readout, and $\alpha$ is an output multiplier.
After RMSNorm with unit gain, $\lVert h \rVert^2 = w$.
If the entries of $W$ are drawn independently with mean zero and variance $\sigma^2$, and $W$ is not tied to the input embedding, the logits of different tokens are independent given $h$, each with mean zero and variance
\begin{equation}\label{eq:logit-variance}
s^2 \;=\; \alpha^2 \sigma^2 w.
\end{equation}
For Gaussian logits, the law of large numbers over the $V$ tokens gives $\frac{1}{V}\sum_{u=1}^{V} e^{z_u} \approx \E\big[e^{z}\big] = e^{s^2/2}$, so the softmax gives token $v$ the probability
\begin{equation}\label{eq:init-softmax}
p(v) \;=\; \frac{e^{z_v}}{\sum_{u=1}^{V} e^{z_u}} \;\approx\; \frac{1}{V}\, e^{\,z_v - s^2/2}.
\end{equation}
Each token gets the uniform probability $1/V$ times a random factor whose mean is one.
The text being scored does not depend on the draw of $W$, so the expected code length of each of its tokens is
\begin{equation}\label{eq:init-code-length}
\E\big[-\log p(t)\big] \;\approx\; \log V + \frac{s^2}{2}.
\end{equation}
The extra $s^2/2$ nats per token is the expected Kullback--Leibler divergence $D_{\mathrm{KL}}(p^{\mathrm{unif}} \,\|\, p)$, and it is zero exactly when $s^2 = 0$.
A simulation of Gaussian logits at $V = 2^{15}$ matches $s^2/2$, for the extra code length and for the reverse divergence $D_{\mathrm{KL}}(p \,\|\, p^{\mathrm{unif}})$, to within $0.2\%$ at five positive values of $s^2$ up to $2$.

\paragraph{$\mu$P and its successors shrink the logit scale with width.}
$\mu$P sets the scale of every layer so that each layer learns features at a rate that does not depend on the width, which lets hyperparameters tuned on a small model transfer to a large one \citep{yang2021tensoriv, yang2021tensorv}.
Its readout makes the logits at initialization vanish as the width grows: with $s_0^2$ the logit variance at the base width $w_0$ where the hyperparameters are tuned,
\begin{equation}\label{eq:mup-logit-variance}
s^2(w) \;=\; s_0^2\, \frac{w_0}{w}.
\end{equation}
\citet{yang2021tensorv} also report better transfer when the readout is initialized at zero; then $s^2 = 0$, and the model is exactly uniform at initialization.
u-$\mu$P \citep{blake2025umup} draws every weight with unit variance, multiplies the readout by $1/w$, and scales the logits by a tuned $\alpha_{\mathrm{loss}}$ before the softmax, so $s^2 = \alpha_{\mathrm{loss}}^2 / w$.
Depth-$\mu$P scales each residual branch by one over the square root of the depth \citep{yang2024tensorvi, bordelon2024depthwise}, and CompleteP by one over the depth \citep{dey2025completep}; both keep the readout of $\mu$P.
In a pre-norm transformer, neither changes $s^2$, because the final normalization fixes $\lVert h \rVert^2 = w$ at every depth.

\paragraph{Example.}
A readout tuned with $s_0^2 = 0.1$ at $w_0 = 256$ has $s^2 = 0.0063$ at width $4{,}096$.
Its expected code length exceeds that of $p^{\mathrm{unif}}$ by $0.0031$ nats, or $0.0045$ bits, per token, against $15$ bits per token for $p^{\mathrm{unif}}$ at $V = 2^{15}$.

\paragraph{The head start remains for every $s^2$.}
The logit variance of \Cref{eq:logit-variance} does not depend on the tokenizer, so by \Cref{eq:init-code-length} a randomly initialized model scores
\begin{equation}\label{eq:init-bpb}
\bpb \;\approx\; \frac{N_\tau}{B} \left( \log V_\tau + \frac{s^2}{2} \right),
\end{equation}
in the nats per byte of \Cref{sec:tokenizer_agnostic}: tokens per byte times a cost per token that still grows with $\log V_\tau$.
The comparison across tokenizers of \Cref{sec:tokenizer_agnostic} therefore keeps a head start for every $s^2$, and $p^{\mathrm{unif}}$ is the case $s^2 = 0$.

\end{document}